\documentclass[11pt,hidelinks]{article}

\usepackage{hyperref}
\usepackage{url}
\usepackage[margin=1.15in]{geometry}
\usepackage{amsmath,amssymb,amsthm, amsfonts}
\usepackage[T1]{fontenc}
\usepackage{mathtools}
\usepackage{xcolor}
\usepackage{enumitem, comment, xifthen}
\usepackage{graphicx}
\graphicspath{{figs/}}
\usepackage{bm}
\usepackage{bbm}
\usepackage{hyperref}
\usepackage{url}
\usepackage{amsmath,amssymb,amsthm, amsfonts}
\usepackage[T1]{fontenc}
\usepackage{mathtools}
\usepackage{xcolor}
\usepackage{enumitem, comment, xifthen}
\usepackage{graphicx}
\graphicspath{{figs/}}
\newcommand\numsamples{\ensuremath{T}}
\newcommand\numseries{\ensuremath{n}}

\newcommand\dimension{\ensuremath{d}}

\let\dim\dimension

\newcommand{\T}{\mathsf{T}}

\newcommand\ARest{\ensuremath{\hat \theta^{\rm AR}}}

\makeatletter
\def\letterdef#1#2#3{\def\letterdef@##1{\expandafter\def\csname #1\endcsname{#2}}%
  \letterdef@@#3{?\@car{}}\@nil}
\def\letterdef@@#1{\@gobble#1\letterdef@{#1}\letterdef@@}
\makeatother

\makeatletter
\newcommand*{\coloneq}{\mathrel{\rlap{%
                     \raisebox{0.3ex}{$\m@th\cdot$}}%
                     \raisebox{-0.3ex}{$\m@th\cdot$}}%
  =}
\makeatother
\makeatletter
\newcommand*{\eqcolon}{=\mathrel{\rlap{%
                     \raisebox{0.3ex}{$\m@th\cdot$}}%
                     \raisebox{-0.3ex}{$\m@th\cdot$}}%
}
\makeatother
\DeclarePairedDelimiterX{\klx}[2]{(}{)}{%
  #1\;\delimsize\|\;#2%
}
\DeclarePairedDelimiterX{\norm}[1]{\|}{\|}{%
  #1%
}

\DeclarePairedDelimiterX{\quantklx}[3]{(}{)}{%
  #1\;\delimsize\|\;#2\;\delimsize\vert\;#3%
}
\DeclarePairedDelimiterX{\inner}[2]{\langle}{\rangle}{%
  #1,#2%
}

\newcommand{\R}{\mathbf R} % reals
\letterdef{c#1}{\mathcal{#1}}{ABCDEFGHIJKLMNOPQRSTUVWXYZ}
\let\defn\coloneq

\newcommand{\floor}[1]{\left\lfloor #1 \right\rfloor}

\newcommand{\e}{\mathrm{e}}

\newcommand{\1}{\mathbf 1} % ones
\let\ones\1
 
\let\epsilon\varepsilon
\newcommand{\eps}{\varepsilon}

\let\hat\widehat
\let\tilde\widetilde

\let\preceq\preccurlyeq

\renewcommand{\leq}{\leqslant}
\renewcommand{\geq}{\geqslant}

\newcommand{\argmin}{\mathop{\rm arg\,min}}

\newcommand{\iid}{\textnormal{i.i.d.}}
\newcommand{\ind}{\textnormal{ind.}}
\newcommand{\simiid}{\stackrel{\iid}{\sim}} % drawn IID
\newcommand{\simind}{\stackrel{\ind}{\sim}} % drawn independently
\makeatletter
\newcommand{\E}{\operatorname*{\mathbf{E}}\ilimits@}
\makeatother
\makeatletter
\renewcommand{\P}{\operatorname*{\mathbf{P}}\ilimits@}
\makeatother

\newcommand{\ie}{\textit{i}.\textit{e}., }

\newcounter{algorithmctr}
\renewcommand{\thealgorithmctr}{\arabic{algorithmctr}}
\newenvironment{algdesc}%
   {\refstepcounter{algorithmctr}\begin{list}{}{%
       \setlength{\rightmargin}{0\linewidth}%
       \setlength{\leftmargin}{0\linewidth}}%
       \rmfamily\small
       \item[]{\setlength{\parskip}{0ex}\hrulefill\par%
        \nopagebreak{\bfseries\textsf{Algorithm \thealgorithmctr~}}}}%
   {{\setlength{\parskip}{-1ex}\nopagebreak\par\hrulefill} \end{list}}

\makeatletter
\long\def\@makecaption#1#2{
        \vskip 0.8ex
        \setbox\@tempboxa\hbox{\small {\bf #1.} #2}
        \parindent 1.5em 
        \dimen0=\hsize
        \advance\dimen0 by -3em
        \ifdim \wd\@tempboxa >\dimen0
                \hbox to \hsize{
                        \parindent 0em
                        \hfil 
                        \parbox{\dimen0}{\def\baselinestretch{0.96}\small
                                {\bf #1.} #2
                                } 
                        \hfil}
        \else \hbox to \hsize{\hfil \box\@tempboxa \hfil}
        \fi
        }
\makeatother

\theoremstyle{plain}

\newtheorem{theo}{Theorem}

\newtheorem{lem}{Lemma}
\newtheorem{prop}{Proposition}
\newtheorem{cor}{Corollary}

\theoremstyle{definition} 

\newtheorem{nota}{Notation}
\newtheorem{de}{Definition}
\newtheorem{exa}{Example}
\newtheorem{as}{Assumption}
\newtheorem{alg}{Algorithm}

\newcommand{\btheo}{\begin{theo}}
\newcommand{\bde}{\begin{de}}
\newcommand{\ble}{\begin{lem}}
\newcommand{\bpr}{\begin{prop}}
\newcommand{\bno}{\begin{nota}}
\newcommand{\bex}{\begin{exa}}
\newcommand{\bcor}{\begin{cor}}
\newcommand{\spro}{\begin{proof}}
\newcommand{\bas}{\begin{as}}
\newcommand{\balg}{\begin{alg}}

\newcommand{\etheo}{\end{theo}}
\newcommand{\ede}{\end{de}}
\newcommand{\ele}{\end{lem}}
\newcommand{\epr}{\end{prop}}
\newcommand{\eno}{\end{nota}}
\newcommand{\eex}{\end{exa}}
\newcommand{\ecor}{\end{cor}}
\newcommand{\fpro}{\end{proof}}
\newcommand{\eas}{\end{as}}
\newcommand{\ealg}{\end{alg}}

\theoremstyle{plain}

\newtheorem{theos}{Theorem}
\newtheorem{props}{Proposition}
\newtheorem{lems}{Lemma}
\newtheorem{cors}{Corollary}

\theoremstyle{definition}
\newtheorem{exas}{Example}
\newtheorem{algs}{Algorithm}
\newtheorem{asss}{Assumption}
\newtheorem{defns}{Definition}

\newcommand{\btheos}{\begin{theos}}
\newcommand{\etheos}{\end{theos}}
\newcommand{\bprops}{\begin{props}}
\newcommand{\eprops}{\end{props}}
\newcommand{\bdes}{\begin{defns}}
\newcommand{\edes}{\end{defns}}
\newcommand{\blems}{\begin{lems}}
\newcommand{\elems}{\end{lems}}
\newcommand{\bcors}{\begin{cors}}
\newcommand{\ecors}{\end{cors}}
\newcommand{\bexs}{\begin{exas}}
\newcommand{\eexs}{\end{exas}}
\newcommand{\balgs}{\begin{algs}}
\newcommand{\ealgs}{\end{algs}}
\newcommand{\bass}{\begin{asss}}
\newcommand{\eass}{\end{asss}}

\usepackage{microtype}
\usepackage{graphicx}
\usepackage{hyperref}
\usepackage{overpic}
\usepackage{subcaption}
\usepackage{booktabs}
\usepackage[square,numbers]{natbib}
\newcommand{\ours}{{\textit{Cluster~\&~Conquer}}}
\newcommand\trueclus[1]{\ensuremath{\mathsf{c}^\star(#1)}}
\newcommand\predclus[1]{\ensuremath{\hat{\mathsf{c}}_{\rm spec}(#1)}}
\newcommand\estclus[1]{\ensuremath{\hat{\mathsf{c}}(#1)}}
\newcommand\anclus[1]{\ensuremath{\hat{\mathsf{c}}_{\rm appx}(#1)}}
\newcommand\matnormsubscript[2]{\left\| #1 \right\|_{#2}}
\newcommand\fronorm[1]{\ensuremath{\matnormsubscript{#1}{\rm F}}}
\newcommand\opnorm[1]{\ensuremath{\matnormsubscript{#1}{\rm op}}}
\newcommand\optcenter{\ensuremath{\theta^\star}}

\newcommand\eigmax{\ensuremath{\gamma}}
\newcommand\numclus{\ensuremath{k}}
\newcommand\numdata{\ensuremath{n}}
\newcommand\OptCenterMat{\ensuremath{\Theta^\star}}
\let\TrueCenterMat\OptCenterMat
\newcommand\EstCenterMat{\ensuremath{\hat \Theta}}
\newcommand\NoiseMat{\ensuremath{G}}
\newcommand\noisevar{\ensuremath{g}}
\newcommand\estcenter{\ensuremath{\hat \theta}}
\newcommand\badindices{\ensuremath{\cI_{\rm bad}}}
\newcommand\goodindices{\ensuremath{\cI_{\rm good}}}
\newcommand\concevent{\mathcal{E}}
\newcommand\VARestoracle{\ensuremath\Gamma^{\rm VAR, or}} 
\newcommand\VARest{\ensuremath\Gamma^{\rm VAR}} 
\def\*#1{\mathbf{#1}}
\def\_#1{\mathcal{#1}}
\def\-#1{\mathbb{#1}}

\usepackage[symbol]{footmisc}

\title{\bf \Large Cluster-and-Conquer: A Framework for Time-Series Forecasting}
\author{Reese Pathak$^1$
\and Rajat Sen$^2$
\and Nikhil Rao$^3$ 
\and N.\ Benjamin Erichson$^{4,5}$
\and Michael I.\ Jordan$^{1, 5}$
\and Inderjit S.\ Dhillon$^{3, 6}$ 
}
\date{{\normalsize 
$^1$Department of Electrical Engineering and Computer Sciences, UC Berkeley\\
$^2$Google Research\\
$^3$Amazon\\
$^4$School of Engineering, University of Pittsburgh\\
$^5$Department of Statistics, UC Berkeley\\
$^6$Department of Computer Science, UT Austin\\[3ex]} 
\today}
\begin{document}

\maketitle 

\begin{abstract}
  We propose a three-stage framework for forecasting high-dimensional time-series data. Our method first estimates parameters for each univariate time series. Next, we use these parameters to cluster the time series. These clusters can be viewed as multivariate time series, for which we then compute parameters. The forecasted values of a single time series can depend on the history of other time series in the same cluster, accounting for intra-cluster similarity while minimizing potential noise in predictions by ignoring inter-cluster effects. Our framework---which we refer to as ``cluster-and-conquer''---is highly general, allowing for any time-series forecasting and clustering method to be used in each step. It is computationally efficient and embarrassingly parallel.%izable. % and typically more efficient than dense vector autoregression. 
  We motivate our framework with a theoretical analysis in an idealized mixed linear regression setting, where we provide guarantees on the quality of the estimates. We accompany these guarantees with experimental results that demonstrate the advantages of our framework: when instantiated with simple linear autoregressive models, we are able to achieve state-of-the-art results on several benchmark datasets, sometimes outperforming deep-learning-based approaches.
\end{abstract}

\section{Introduction}
\label{sec:intro}
High-dimensional time-series forecasting is crucial in applications such as finance~\cite{zhu2002statstream}, e-commerce~\cite{faloutsos2020forecasting}, medical data analysis~\cite{matsubara2014funnel}, fluid flows and climatology~\cite{divine2012climate,azencot2020forecasting}. Modern time-series datasets can have millions of correlated time series; e.g., demand forecasting for an online store like Amazon might require predicting the future sales of millions of items, and these demands are generally inter-related. Thus time-series forecasting methods should be scalable and equipped with the ability to handle inter-time-series correlation.

Most time-series forecasting methods operate at the two extremes: (1) \textbf{Univariate} modeling where the future prediction of a time series depends on its own past and associated covariates; (2) \textbf{Multivariate} modeling where the future prediction of a time series can potentially depend on the past of \textit{all} the time series in the dataset. 
Univariate models are %typically 
unable to capture inter time-series correlations during predictions, yet are computationally efficient and/or easy to parallelize. 
On the other hand, multivariate time-series models  
are better able to capture these time-series correlations, yet can suffer from computational inefficiencies. We give a more extensive comparison of prior work within these two extremes, in Section~\ref{sec:rwork}. 

The disadvantages of univariate and multivariate time-series modeling point to a need to come up with forecasting methods that can achieve a middle ground, i.e, the future values of a time series are modeled as a function of its own past and only a few other related time series. Therefore, in this paper we ask the following question: \emph{For high-dimensional time-series problems, is it possible to develop statistically and computationally efficient methods that exploit only relevant inter-time-series relationships?}. Prior works like~\citep{bandara2020forecasting} have indeed shown some promising empirical results in this direction, but detailed models of this problem with theoretical grounding is still lacking.

In this paper, we tackle this question in both theory and practice, and answer it in the affirmative. We propose a `cluster-and-conquer' framework. Given $n$ time series, we first use highly scalable univariate models (such as scalar autoregression) to learn parameters of each time series. Then, we apply efficient clustering algorithms on these parameters, to partition the collection of time series into $k$ clusters. Finally, for each cluster, we apply a joint forecasting approach (using all the time series only in the cluster). 

The \textit{main contributions} of this paper are a framework for time-series modelling with the following properties:
\begin{itemize}
    \item [(i)] By varying the number of clusters, 
    we are able to interpolate between univariate and multivariate modelling, and reap the statistical and computational benefits of each paradigm. 
    \item [(ii)]
    When employed with standard linear autoregressive models, our method shows strong empirical performance. 
    Specifically, we are able to outperform 
    many previously proposed 
    time-series modelling approaches over a 
    wide variety of datasets.  
    \item [(iii)]
    Our framework---when analyzed in a mixture-of-regressions setting---enjoys theoretical guarantees. Namely, we can show that this framework is able 
    to ensure recovery of model parameters at standard 
    parametric rates.
\end{itemize}

\textbf{Notation.} A multivariate time-series dataset with $n$ time series and $T$ time points can be represented as $\*X \in \R^{n \times T}$. For integers, $i \leq j$, $i:j := \{i, i+1, \cdots, j\}$. For a positive integer $n$, $[n]$ denotes the set $\{1, 2, \cdots, n\}$. The notation $x_{i}^{(\cJ)}$ denotes the values at the time points in $\cJ$ for time series $i$ represented as a column vector, in ascending order of the indices in $\cJ$. $\norm{\cdot}_p$ denotes the $\ell_p$ norm for vectors. The operator and Frobenius norms of matrices are denoted by $\opnorm{\cdot}$ and $\fronorm{\cdot}$ respectively. For two vectors $u \in \R^p, v \in \R^q$, we write $u \otimes v$ for their tensor (outer) product, which is  a $p \times q$ matrix with entries $u_i v_j$, $i \in [p], j \in [q]$. 

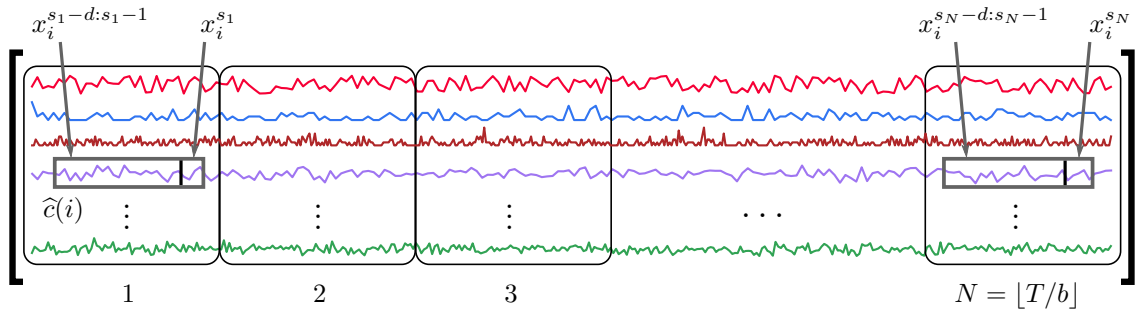
\begin{figure*}[!h]\vspace{+0.5cm}
	\centering
	\begin{overpic}[width=0.95\textwidth]{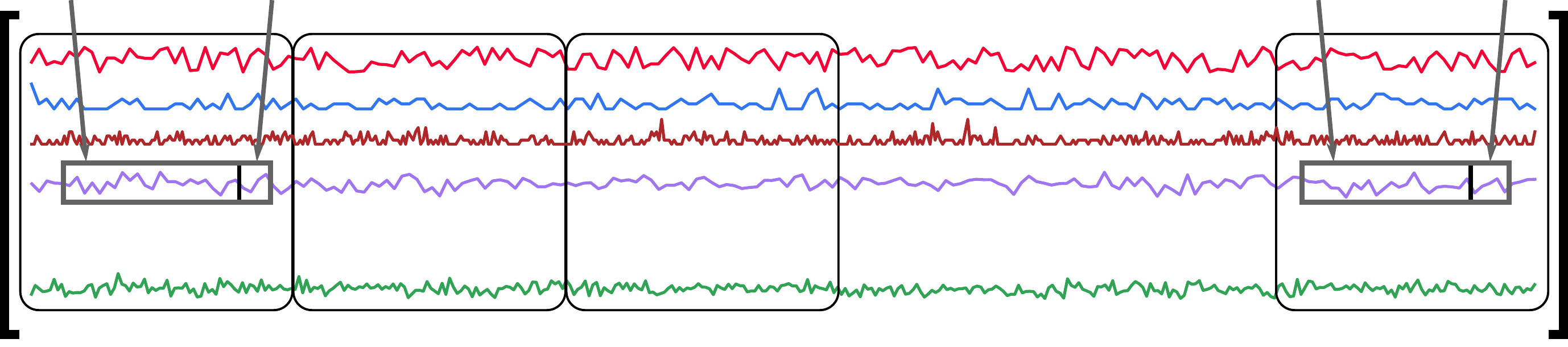} 
		%\put(-6,15){\rotatebox{90}{\footnotesize test accuracy}}			
		\put(10,-1){\small {$1$}} 
		\put(27,-1){\small {$2$}} 
		\put(44,-1){\small {$3$}} 
		\put(84,-1){\small {$N=\floor{T/b}$}} 
		
		\put(96,23){\small {$x_i^{s_N}$}} 
		\put(80.8,23){\small {$x_i^{s_N-d:s_N-1}$}} 	
		
		\put(17,23){\small {$x_i^{s_1}$}} 
		\put(2,23){\small {$x_i^{s_1-d:s_1-1}$}} 		 
		
		\put(3,6.5){\small {$\hat{c}(i)$}} 		 
		
		\put(10,5.5){\Large {$\vdots$}} 
		\put(27,5.5){\Large {$\vdots$}} 		 		 	
		\put(44,5.5){\Large {$\vdots$}} 
		\put(89,5.5){\Large {$\vdots$}}	 
		\put(65,6.5){\Large {$\hdots$}}			 		 			 		 			 		 	
	\end{overpic}\vspace{+0.0cm}		
	
	\caption{\small Illustration of forming regression datasets from time-series data blocks. Since $b$ is a multiple of the mixing time of the time series, the samples for the regression problem will be approximately be independent and identically distributed.
	Above, $\{s_i\}_{i=1}^N$ are the sample time points selected within each chunk.}
	\label{fig:reg}
\end{figure*}

\section{Related Work}
\label{sec:rwork}
There is a vast body of work in classical time-series forecasting;  we refer the reader to~\citet{hyndman2018forecasting} for an overview. Here, we mainly focus on the time-series forecasting literature in the machine learning community; see also \citet{benidis2020neural}.

\paragraph{Univariate modeling.} Basic forecasting techniques train one model per dataset but the prediction using the trained model is done for each time series and its covariates individually.
Specifically, in univariate modeling, we typically use predictions of the form 
\[
\hat x_{i}^{(T + 1)}
= \hat f_i(x_{i}^{(1:T)}), \quad 
i = 1,\dots, n.\footnote{We use $\hat x, \hat f$ to denote estimates.}
\]
Autoregressive (AR) models are a classic example that are computationally scalable and enjoy low variance, but these models typically suffer from high statistical bias in that they cannot account for complicated inter-time-series relationships. 
In contrast, recent deep learning methods~\cite{salinas2019deepar, borovykh2017conditional,lipschitzrnn,lim2021noisy, rangapuram2018deep} aim to address this problem by training shared weights of a large model on the whole dataset---effectively selecting $\hat f_i$ based on the other time series 
$\{x^{(t)}_{i}\}_{t \leq T, j \neq i}$.  At 
prediction time the estimate of future values of each univariate time series is modelled by a function of its own past and covariates. 

\paragraph{Multivariate modeling.} More advanced forecasting techniques aim to capture dependencies across different time series in the dataset. Prediciton for each time series is computed as a function of 
the history of \emph{all} the other time series. 
Specifically, in multivariate modeling, we typically use predictions of the form
\[
\hat x_{i}^{(T + 1)} = 
\hat f_i(x_{1}^{(1:T)}, , 
x_{2}^{(1:T)},  \dots, x_{n}^{(1:T)}).
\]

Models that use this paradigm include dense vector autoregression (VAR), 
dynamic linear models (DLMs), TRMF~\cite{yu2016temporal}, DeepGLO~\cite{sen2019think} and graph neural network based models~\cite{li2018diffusion, cao2020spectral, yu2018spatio, wu2019graph, guo2016high, bai2020adaptive,erichson2019physics}. However, some of these models are computationally inefficient. For instance, VAR does not scale beyond even moderately sized datasets. 
The complexity to fit a VAR model on a dataset of $n$ time series scales as $O(n^6)$ \cite{yu2016temporal}. Similarly, DLMs typically scale as $O(n^3)$, which is again prohibitive for large-$n$ problems. Additionally, these models typically have a large number of parameters, and can therefore suffer from high model variance. At prediction time, forecasts can be made using contributions from potentially irrelevant time series which can lead to higher statistical error.

\paragraph{Time-series clustering.} 

The vast literature on time-series clustering is also relevant to our work. We refer the reader to the surveys by~\citet{liao2005clustering} and \citet{aghabozorgi2015time} for an in-depth treatment. Model-based clustering techniques that assume that similar time series are governed by similar model parameters are especially relevant to our work~\cite{kalpakis2001distance, ding2015yading, kini2013bayesian}. In recent work~\cite{bandara2020forecasting}, Bandara and co-authors consider a similar cluster-based time-series prediction
framework. However, there are two main 
differences: first, the authors do not present theory to justify or accompany their modelling framework. Secondly, their within-cluster model does not forecast individual time series based on the entire history of the other time series, and 
could miss important inter-cluster time-series correlations.
\section{Cluster-and-Conquer Framework}
\label{sec:framework}

Suppose we are given a history of $T$ timepoints of 
$n$ univariate time series, $\{x_{i}^{(t)}\}_{i \leq n, t \leq T}$. 
We would like to produce a prediction, for the next time point, which is at time $T + 1$.\footnote{For simplicity we write all our equations for predicting one time point in the future, however all the ideas trivially extend to multi-step forecasting.} We want our predictions to satisfy  $\hat x_{i}^{(T + 1)} \approx x_{i}^{(T + 1)}$, 
for all time series $i = 1, \dots, n$. 
Now, to overcome some of the issues discussed in the related work section, we propose a three-stage procedure for time-series forecasting.

Our framework consists of the following three steps:
\begin{itemize}
\item {\bf Local step:~} We first divide the time-series dataset into $N = \floor{T/b}$ chunks of contiguous times as shown in Figure~\ref{fig:reg}; within each of the $N$ chunks, we select a time point $s_i, i \in [N]$, for which we sample the data. At a high level, we can think of selecting $b$ as a multiple of the (largest) mixing time of the time series. We can now extract one sample for a local autoregressive regression problem per time series from each block. For example for the $i$-th time series the sample from the $j$-th block can be 
\[
(\text{covariates}, \text{target}) = (x^{(s_j-d:s_j-1)}_{i}, x_{i}^{(s_j)}).
\]
For each time series we collect all the samples and solve a least-squares problem, and hence this step is efficient. The AR parameters obtained are $\{\ARest_i\}_{i=1}^{n}$.   

\item {\bf Clustering step:~} Next, we cluster the parameters of the AR model. We can use a clustering algorithm, for instance spectral clustering~\cite{kannan2009spectral}, or METIS~\cite{kalpakis2001distance} with inputs $\{\ARest_i\}_{i=1}^{n}$ to obtain a clustering of time series into $k$ clusters. Let $\estclus{i}$ be the estimated cluster identity of time series $i$.

\item {\bf Global step:~} Given a partition, we can estimate VAR parameters for every cluster. We can reuse the block structure from above and extract one sample per block for the VAR problem associated with each time series. For instance, for the $i$-th time series the sample for the $j$-th block can be 
\begin{align*}
	\text{covariates} & = x^{(s_j-d:s_j-1)}_{i}, \{x^{(s_j-d:s_j-1)}_{l}\}_{\estclus{l} = \estclus{i}}, \\
	\text{target}  &=  x^{(s_j)}_{i}.
\end{align*}
The covariates are the history of the $i$th time series, together with all the histories of the time series assigned 
to the same cluster---that is, time series $l$ satisfying 
$\estclus{l} = \estclus{i}$.
Therefore, in this model the future of a time series can depend on the past of all time series in the same cluster, through the learned VAR coefficients. Note that we use VAR here for exposition and to make it amenable to theoretical analysis. In practice, this step can be solved using more complex low-rank \cite{yu2016temporal} or deep-learning \cite{sen2019think} approaches.
\end{itemize}

Each of the samples for the two regression problems above should be close to i.i.d as $b$ is a multiple of the mixing time of the processes. Therefore we will analyze an i.i.d analogue of the above system which is similar to a mixed linear regression problem. The mixed linear regression problem is a useful surrogate of our problem in an i.i.d setting that lets us analyze the salient features of our framework---parameter recovery of local AR models, recovery of clusters using clustering on the recovered AR parameters and recovery of VAR parameters within cluster---while avoiding the complexities arising from non-i.i.d data. Note that risk bounds and parameter recovery bounds for simple AR models are still an area of active research~\cite{goldenshluger2001nonasymptotic, mcdonald2017nonparametric, kuznetsov2015learning} and we would like to decouple the hardness of analyzing our framework from the issues arising from a non-i.i.d dataset. These issues can be carefully dealt with using tools from mixing time literature and are left for future work. Analogous i.i.d settings have been used before in the literature to analyze interesting properties of time-series clustering algorithms~\cite{kini2013bayesian, ding2015yading}.

\section{Theoretical Results}
\label{sec:theory}
In this section we provide our formal algorithm and associated parameter recovery guarantees in the Mixed Linear Regression (MLR) setting. 
Although idealized, this formal model allows us to state formal guarantees about parameter recovery. Additionally, we refer the reader to 
Appendix~\ref{sec:lgn}, where we further explain how this setting 
is closely related to the original time-series problem.

{\bf MLR model: } We consider the following fixed design mixed linear regression problem. 
There are $n$ regression designs each with $T$ samples. $x_i^{(t)}$ denotes the $t$-th covariate from the $i$-th design. We will use the notation $X^{(t)} = \{x_i^{(t)}\}_{i=1}^n$, where $X^{(t)} \in \R^{\numseries \times \dim}$.

We assume that for some positive integer $k$, we have 
(unknown) cluster labels, $\trueclus{1}, \dots, \trueclus{\numseries} \in [k]^n$, for each regression problem, where for all $i \in [\numseries], t \in [\numsamples]$:
\[
\theta_i \simind \mathsf{N}_d(\overline{\theta}_{\trueclus{i}}, \nu^2 I_d).
\]
We then observe $\{y_i^{(t)}\}_{i=1}^n$ for $t = 1, \dots, T$
according to the following linear observational model:
\begin{equation}\label{eqn:observation-model}
	y_i^{(t)} = \theta_i^\T x_i^{(t)} + \sum_{\substack{j \neq i  \\ \trueclus{j} = \trueclus{i}}} \gamma_{ij}^\T x_j^{(t)} + \epsilon_i^{(t)}, \quad \mbox{for}~
	i \in [\numdata].
\end{equation}
For simplicity, we assume that 
\[
\eps_i^{(t)} \simiid \mathsf{N}(0, \sigma^2) \quad \mbox{and} \quad 
\gamma_{ij} \simiid \mathsf{N}_d(0, \tau^2 I_d).
\]

The target $y_i^{(t)}$ can be thought of as the future value of $i$th time series that needs to be predicted. It depends linearly on its own past $x_i^{(t)}$ and also the past of the other time series in the same cluster $\{x_j^{(t)}\}$. The local AR parameters $\theta_i$ for each cluster come from the same Gaussian distribution.

\newcommand{\EmpCov}{\hat \Sigma}
\subsection{Algorithm for Mixed Linear Regression}
\label{sec:iid_algorithm}
Our framework for the mixed linear regression problem is instantiated as Algorithm~\ref{alg:mlr}. 

\begin{algdesc}
\emph{\small \ours~for Mixed Linear Regression.}
\label{alg:mlr}
\begin{tabbing}
  {\bf input:} $y_i^{(t)}, x_i^{(t)}$ for $i \in [n], t \in [T].$ \\[1.4ex]
  \quad \=\ 1. \emph{Train one AR model per time series $\{\hat{\theta}_i^{(AR)}\}_{i=1}^{n}$,} \\
  \quad \=\ \emph{according to Equation~\eqref{prob:scalar-AR}.}\\[1.7ex]
  \quad \=\ 2. \emph{Estimate cluster indices $\{\estclus{i}\}_{i=1}^{n}$ using Algorithm~\ref{alg:spectral-clustering}.}\\[1.3ex]
  \quad \=\ 3. \emph{Solve VAR models for each cluster as in~\eqref{eq:var}}.
\end{tabbing}
\end{algdesc}

\subsection{Overview of Guarantees} 
We now analyze Algorithm~\ref{alg:mlr} in the mixed linear regression setting defined in  Section~\ref{sec:theory}, under the following assumptions. 
\bas\label{as:clus-ratio}
The number of clusters $k \geq \rho n$ for
some constant $\rho \in (0, 1)$. 
\eas
\bas\label{as:clus-balanced}
The cluster sizes are uniform: each cluster has $m = \numseries/k$
time series associated to it. 
\eas 
\bas\label{as:isotropy}
Within each cluster, the design is isotropic. 
More formally, for cluster $\iota \in [\numclus]$, 
suppose that time series $i_1, \dots, i_m \in [\numseries]$ 
belong to this cluster (\ie $\trueclus{i_j} = \iota$ 
for $j \in [m]$). Define for each $t \in [\numsamples]$
\[
x^{(t)}_\iota = 
(x_{i_1}^{(t)}, \dots, x_{i_m}^{(t)}) 
\in \R^{m\dimension}
\]
Then we assume
$
\frac{1}{\numsamples} \sum_{t=1}^\T x_\iota^{(t)}  \otimes x_\iota^{(t)} 
= I_{md}, ~~ \mbox{for all}~\iota \in [\numclus]. 
$
\eas 

Our analysis will aim to demonstrate that Algorithm~\ref{alg:mlr}
provides good estimates of the autoregression parameters. 
To do so, we begin in Section~\ref{sec:scalar-AR} by 
providing a stochastic decomposition for 
the scalar AR parameters estimated for each time series.
Having done so, we will establish that the scalar
AR parameters are nearly a Gaussian mixture model. 
We are then able to extend existing guarantees for Gaussian 
mixtures to our setting, giving explicit conditions under 
which we can exactly recover the cluster labels 
for the mixed linear regression model via spectral 
clustering on the scalar AR estimates. 
These guarantees are provided in Section~\ref{sec:clustering}. 
We then study the properties of a single cluster-wide 
vector autoregression, and provide 
a high probability bound on the errors incurred 
from model noise. By union bounding over the 
clusters and the failure probability in our 
exact clustering guarantee, we are able to 
give a formal statement on the recovery guarantees 
of Algorithm~\ref{alg:mlr}.

\subsection{Scalar Autoregression} 
\label{sec:scalar-AR}
As the first step of our procedure we propose fitting scalar autoregression parameter estimates according to a $\mathrm{AR}(d)$ model. 
Formally, we estimate $\ARest_i \in \R^d$, for $i \in [\numseries]$ 
via the following empirical risk minimization problem, 
\begin{equation}
\label{prob:scalar-AR}
\ARest_i \defn \argmin_{\theta \in \R^d} \bigg\{\frac{1}{\numsamples} \sum_{t = 1}^{\numsamples} 
\big(\theta^\T x_i^{(t)} - y_i^{(t)}\big)^2\bigg\}.
\end{equation}
Our analysis begins with a stochastic decomposition of 
$\ARest_i$. To state it, we introduce the following notation for the
(weighted) empirical covariance 
matrix, for a given time series. 
For any weights $\rho = (\rho^{(1)}, \dots, \rho^{(\numsamples)}) \in \R^\numsamples$, 
we define
\[
\EmpCov_i(\rho) \defn \frac{1}{T}\sum_{t=1}\T 
(\rho^{(t)})^2 x_i^{(t)} \otimes x_i^{(t)}.
\]
In the special case $\rho = \1$, we write $\EmpCov_i \defn \EmpCov_i(\1).$
With this notation in hand, we can now state the decomposition 
mentioned above. 
\ble
\label{lem:ar-decomp}
For each $i \in [\numseries]$, we have $\ARest_i  \simiid \mathsf{N}_d(\overline{\theta}_{\trueclus{i}}, \Lambda_i)$, 
where 
\[
\Lambda_i \defn \nu^2 I_d + \frac{\sigma^2}{T} \EmpCov_i^{-1} + 
\frac{\tau^2}{T} \EmpCov_i^{-2}\EmpCov_i(\rho_i),
\]
where $\rho_i = (\rho_i^{(1)}, \dots, \rho_i^{(\numsamples)})$, with \\ 
$\rho_i^{(t)} \defn \sqrt{\sum_{j : \trueclus{j} = \trueclus{i}} \|x_j^{(t)}\|_2^2}$, for 
$t \in [\numsamples]$.
\ele 
Whereas the true AR parameters $\theta_i$ follow 
a Gaussian mixture model, 
Lemma~\ref{lem:ar-decomp} shows that $\ARest_i$, while 
not a Gaussian mixture due to the different covariances $\Lambda_i$, 
is at least centered at the cluster mean $\overline \theta_{\trueclus{i}}$ for 
each $i \in [\numseries]$. The proof has been deferred to the appendix in Section~\ref{sec:others}.

We note that under Assumption~\ref{as:isotropy}, the 
covariances above can be further simplified to 
\begin{equation}\label{eqn:assumed-cov}
\Lambda_i \defn 
\Big(\nu^2 + \frac{\sigma^2}{T}\Big)I_d 
+ \frac{\tau^2}{T} \EmpCov_i(\rho_i). 
\end{equation} 
This will be a useful identity in the sequel. 

\subsection{A Result for Near Mixtures of Gaussians}
\label{sec:clustering}
As we demonstrated in Lemma~\ref{lem:ar-decomp}, after 
computing autoregression estimates $\ARest_i$, 
we obtain a near mixture of Gaussians. In order 
to obtain guarantees for the vector autoregression, 
we first show that the clustering estimates 
are consistent. We begin with the following general result for 
near mixtures of Gaussians. 

\paragraph{Setting.} To state the result, we begin by formally defining our setting. 
Suppose there are $\numclus$ clusters, and $\numdata$ data points $X_1, \dots, X_n \in \R^\dimension$.
These data are associated to unknown cluster labels $\trueclus{i}$, where
we view $\mathsf{c}^\star \colon [\numdata] \to [\numclus]$ as the labelling function.
We assume that there are unknown cluster centers,
$\optcenter_1, \dots, \optcenter_\numclus \in \R^\dimension$.
Further assume we observe
\[
X_i \simind \mathsf{N}_\dimension(\optcenter_{\trueclus{i}}, \Sigma_i),
\quad
i = 1, \ldots, \numdata.
\]
We additionally assume that the sample covariances $\Sigma_1, \dots, \Sigma_\numdata$ are unknown.
Note that this is not quite a mixture of Gaussians, since we allow that $\Sigma_i \neq \Sigma_j$,
even if $\trueclus{i} = \trueclus{j}$.

\paragraph{Clustering near Gaussian mixtures.} 
We study the spectral clustering algorithm, Algorithm~\ref{alg:spectral-clustering} and provide the following recovery guarantee for near mixture of Gaussians.

\btheo
\label{thm:main-GMM-result}
Suppose that $\Sigma_i \preceq \eigmax I_d$, for $i \in [\numdata]$.
Suppose Assumption~\ref{as:clus-ratio} holds. 
Assume that the centers are well separated:
\[
\min_{i \neq j \in [k]} 
\|\optcenter_i - \optcenter_j\|_2 \geq 32  \sqrt{\eigmax} k \sqrt{\frac{1 + \tfrac{\dimension}{\numdata}}{\beta}}
\max\left\{1, \frac{\beta}{\rho} \right\}.
\]
Then, with probability at least $1 - \exp(-0.08 \numdata)$,
there exists a permutation $\phi \colon [\numclus] \to [\numclus]$ such that
$\predclus{i} = \phi(\trueclus{i})$ for all $i \in [\numdata]$.
\etheo
The proof is presented in Section~\ref{sec:thm-proof} of the appendix. 
The above theorem implies the following corollary for our mixture of regression problem. 

\bcor
\label{cor:exact-recovery}
Suppose that Assumptions~\ref{as:clus-ratio},~\ref{as:clus-balanced},
and \ref{as:isotropy} hold.
Let
\[
\lambda \defn \nu^2 + \frac{\sigma^2 + \tfrac{\tau^2}{\rho}}{T}. 
\]
If the following separation condition holds,
\[
\min_{i \neq j \in [\numclus]} 
\|\overline{\theta}_{i} - \overline{\theta}_{j}
\|_2
\
\geq  
32  \sqrt{\lambda} k \sqrt{\frac{1 + \tfrac{\dimension}{\numseries}}{\beta}}
\max\left\{1, \frac{\beta}{\rho} \right\},
\]
then spectral clustering on $\{\ARest_i\}_{i=1}^\numseries$ produces
an estimate $\hat{\mathsf{c}}_{\rm spec}$ which is 
exactly equal to the true clustering $\mathsf{c}^\star$ (up 
to relabellings of $\{1, \dots, k\}$) with probability
$1 - \exp(-0.08 \numseries)$. 
\ecor

\begin{algdesc}[Spectral clustering]
\label{alg:spectral-clustering}
\emph{Algorithm for estimating clusters from samples~\cite{kannan2009spectral}.}
\begin{tabbing}
  {\bf input:}\quad sample matrix $X \in \R^{\dimension \times \numdata}$, number of clusters, $\numclus$ \\[1.4ex]
  \quad \=\ 1. \emph{Compute truncated SVD of sample matrix.}\\[1.3ex]
  \quad \=\ \qquad \qquad
  $X_\numclus \defn \sum_{i=1}^\numclus \hat \sigma_i \hat u_i \hat v_i^\T \eqcolon \hat U \hat \Sigma \hat V^\T \in \R^{\dimension \times \numdata}$\\[1.3ex]
  \quad \=\ 2. \emph{Compute projection onto top-$k$ left singular vectors.}\\[1.3ex]
  \quad \=\ \qquad \qquad
  $Y \defn \hat U^\T X_\numclus = \hat \Sigma \hat V^\T \in \R^{\numclus \times \numdata}$ \\[1.3ex]
  \quad \=\ 3. \emph{Solve $\numclus$-means on the columns of $Y$.} Solve:
  \\[1.3ex]
  \quad \=\ \qquad \qquad $\mbox{minimize} \quad \frac{1}{2n} \sum_{i=1}^\numdata \|Y_i - \eta_{\kappa_i}\|_2^2$, \\[1.3ex]
  \quad \=\ \quad with variables $\eta_1, \dots, \eta_\numclus \in \R^\numclus$ and
  $\kappa_1, \dots, \kappa_n \in [k]$.\\[1.4ex]
  {\bf output:}\quad
  labelling $\hat{\mathsf{c}}_{\rm spec} \colon [\numdata] \to [\numclus]$,
  where $\predclus{i} \defn \kappa_i^\star$.
\end{tabbing}
\end{algdesc}

\subsection{Vector Autoregression per Cluster}
\label{sec:vector-AR}
The final step of our argument requires us to compute 
a high-probability guarantee on the closeness of 
estimated vector autoregression parameters. 

For each cluster $\iota \in [\numclus]$, under 
Assumption~\ref{as:clus-balanced}, there is an index set,
\[
\{j \in [n] : \mathsf{c}^\star(j) = \iota\}= \Big\{i_1, \dots, i_m\Big\} 
\subset [\numseries],
\]
that specifies the indices of the time series that truly belong to cluster $\iota \in [\numclus]$. 
Define the following vectorized data: 
\[
y_\iota^{(t)} = \begin{bmatrix} y_{i_1}^{(t)} 
\\ \vdots \\ y_{i_m}^{(t)}
\end{bmatrix} \quad \mbox{and} \quad 
x_\iota^{(t)} = \begin{bmatrix} x_{i_1}^{(t)} \\ 
\vdots \\ x_{i_m}^{(t)} \end{bmatrix}. 
\]
Hence, $y_\iota^{(t)} \in \R^{m}$ and 
$x_{\iota}^{(t)} \in \R^{m\dimension}$, 
for $\iota \in [\numclus]$ and $t \in [\numsamples]$. 

With this notation in hand, for each cluster $\iota \in [\numclus]$,
the observational model stated in~\eqref{eqn:observation-model} 
is equivalent to 
\[
y_\iota^{(t)} = (I_{m} \otimes (x_\iota^{(t)})^\T) \Gamma_\iota^\star + \varepsilon_\iota^{(t)}, 
\mbox{for all}~t \in [\numsamples],~\iota \in [\numclus]. 
\]
Here we define $\Gamma_\iota^\star \in \R^{m^2 \dimension}$
to be the concatenation of the parameters in cluster $\iota \in [\numclus]$. Specifically, in view of our observational model~\eqref{eqn:observation-model}, suppose that we define 
\[
\tilde \gamma_{i_u i_v} 
\defn 
\begin{cases} 
 \theta_{i_u} & u = v \\ 
\gamma_{i_u i_v} & u \neq v
\end{cases},
\quad \mbox{for}~u, v \in [m].
\]
Then 
\[
\Gamma_\iota^\star = 
\Big(
\tilde \gamma_{i_1 i_1} , 
\tilde \gamma_{i_1, i_2}, 
\ldots, 
\tilde \gamma_{i_2 i_1}, 
\tilde \gamma_{i_2, i_2}, 
\ldots, \gamma_{i_m i_m} \Big)
\in \R^{m^2 d}.
\]
This is our target of estimation. 
\paragraph{Oracle VAR estimator.} 
The first estimator we study is what 
we refer to as the \emph{oracle VAR estimate for cluster~$\iota$}:
\begin{equation}
\scalebox{0.85}{$\VARestoracle_\iota \defn \underset{\Gamma \in \R^{m^2 d}}{\argmin} 
\bigg\{\frac{1}{2T} \sum_{t = 1}^\numsamples 
\|y_\iota^{(t)} - (I_m \otimes (x_\iota^{(t)})^\T) \Gamma\|_2^2
\bigg\}.$}
\end{equation}
Crucially, it assumes knowledge of the set $\{i_1, \dots, 
i_m\}$, which are those time series which truly belong to cluster 
$\iota \in [\numclus]$. 

For this estimator, we have the following guarantee. 

\ble
\label{lem:VAR-guarantee} 
Suppose that Assumption~\ref{as:isotropy} holds. 
Then for each $\iota \in [\numclus]$, with probability at least $1 - \delta$ we have
\[
\big\|\VARestoracle_\iota - \Gamma^\star_\iota \big\|_2 
\leq \sqrt{2} \sqrt{\frac{\sigma^2 m^2d}{T}} 
+ \sqrt{3 \frac{\sigma^2}{T} \log \frac{1}{\delta}}. 
\]
\ele 

Deferring the proof to the appendix, an outline is as follows.
First we demonstrate that under the stated 
assumptions,
$
\VARestoracle_i \sim \mathsf{N}(\Gamma^\star, 
\tfrac{\sigma^2}{T} I_{m^2 d})$.
Consequently, 
\[
\E\Big[\|\VARestoracle_i - \Gamma_\iota^\star\|_2^2 \Big] 
\leq \frac{\sigma^2 m^2 d}{T}.
\]
Establishing concentration around the mean 
is a consequence of tail-bounds for $\chi^2$ random 
variates. We provide the full proof in Section~\ref{sec:others} of the appendix.

\paragraph{VAR estimator with estimated clustering.} 

We will show that the preceding result and Corollary~\ref{cor:exact-recovery},
immediately imply our main recovery guarantee, via a simple appeal
to a union bound. To state this result, we first need to 
describe how to construct the VAR estimate when the ground truth 
clustering is unknown. To do this, we define the following
index set:
\[
\{j \in [n] : \estclus{j} = \iota\} = 
\{\hat i_1, \dots, \hat i_{m_\iota}\} \subset [\numseries].
\]
This is the set of time series which are estimated to belong to cluster $\iota$. 
Define 
\[
\hat y_\iota^{(t)} = \begin{bmatrix} y_{\hat i_1}^{(t)} 
\\ \vdots \\ y_{\hat i_{m_\iota}}^{(t)}
\end{bmatrix} \quad \mbox{and} \quad 
\hat x_\iota^{(t)} = \begin{bmatrix} x_{\hat i_1}^{(t)} \\ 
\vdots \\ x_{\hat i_{m_\iota}}^{(t)} \end{bmatrix}. 
\]
Hence, $\hat y_\iota^{(t)} \in \R^{m_\iota}$ and 
$\hat x_{\iota}^{(t)} \in \R^{m_\iota \dimension}$, 
for $\iota \in [\numclus]$ and $t \in [\numsamples]$. 
We now form our \emph{VAR estimate for cluster~$\iota$}:
\begin{equation}
\scalebox{0.85}{$\VARest_\iota \defn \underset{\Gamma \in \R^{m_\iota^2 d}} {\argmin}
\bigg\{\frac{1}{2T} \sum_{t = 1}^\numsamples 
\|\hat y_\iota^{(t)} - (I_m \otimes (\hat x_\iota^{(t)})^\T) \Gamma\|_2^2
\bigg\}.$} \label{eq:var}
\end{equation}

\btheo[Recovery guarantee for mixed linear regression]
Suppose that Assumptions~\ref{as:clus-ratio},~\ref{as:clus-balanced},
and \ref{as:isotropy} hold for the mixed linear 
regression problem. Additionally assume that the separation conditions of 
Corollary~\ref{cor:exact-recovery} hold. Then,
with probability at least 
$1 - \delta - \exp(-0.08\numseries)$, we have the following guarantee:
\[
\max_{\iota \in [\numclus]}
\big\|\VARest_\iota - \Gamma^\star_\iota \big\|_2 
\leq
\sqrt{2} \sqrt{\frac{\sigma^2 m^2d}{T}} 
+ \sqrt{3 \frac{\sigma^2}{T} \log \frac{\numclus}{\delta}}. 
\]
\etheo 

This result is interesting because it indicates that, up to logarithmic and 
constant factors, 
the worst-case recovery error of $\Gamma^\star$ is $\sqrt{\tfrac{\sigma^2 m^2 d}{T}}$, 
which is the standard rate for a $(m^2 d)$-dimensional parameter, with $T$ samples 
and noise variance $\sigma^2$.

\section{Empirical Results}
\label{sec:experiments}

\begin{figure*}[]
	\centering
	%\begin{subfigure}[t]{0.44\textwidth}
	\begin{subfigure}[t]{0.33\textwidth}
		\centering
		\begin{overpic}[width=1\textwidth]{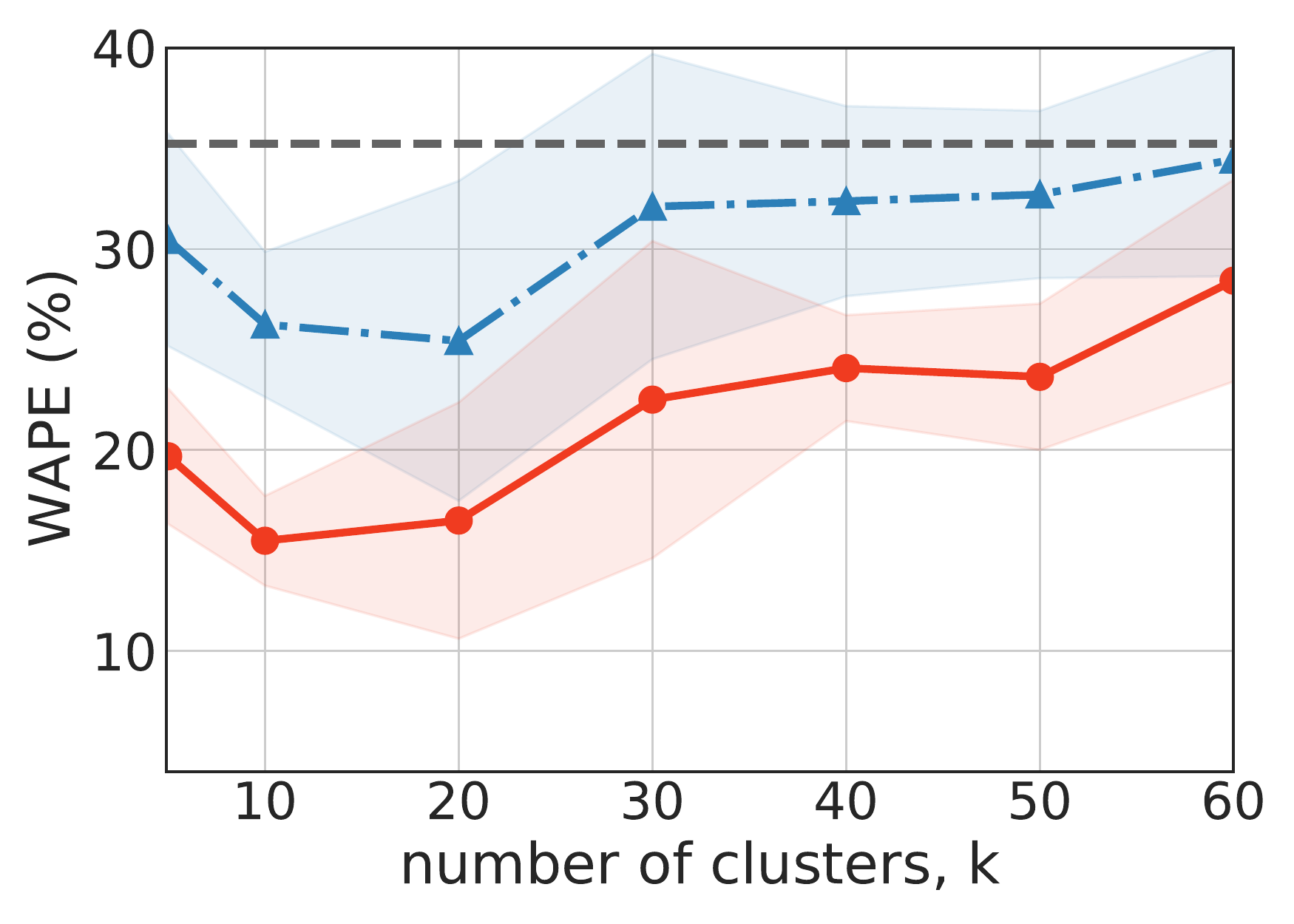}
		\end{overpic}
		%\caption{}
		%\label{fig:k10}
	\end{subfigure} \hspace{-0.6cm}
	~
	\begin{subfigure}[t]{0.33\textwidth}
		\centering
		\begin{overpic}[width=1\textwidth]{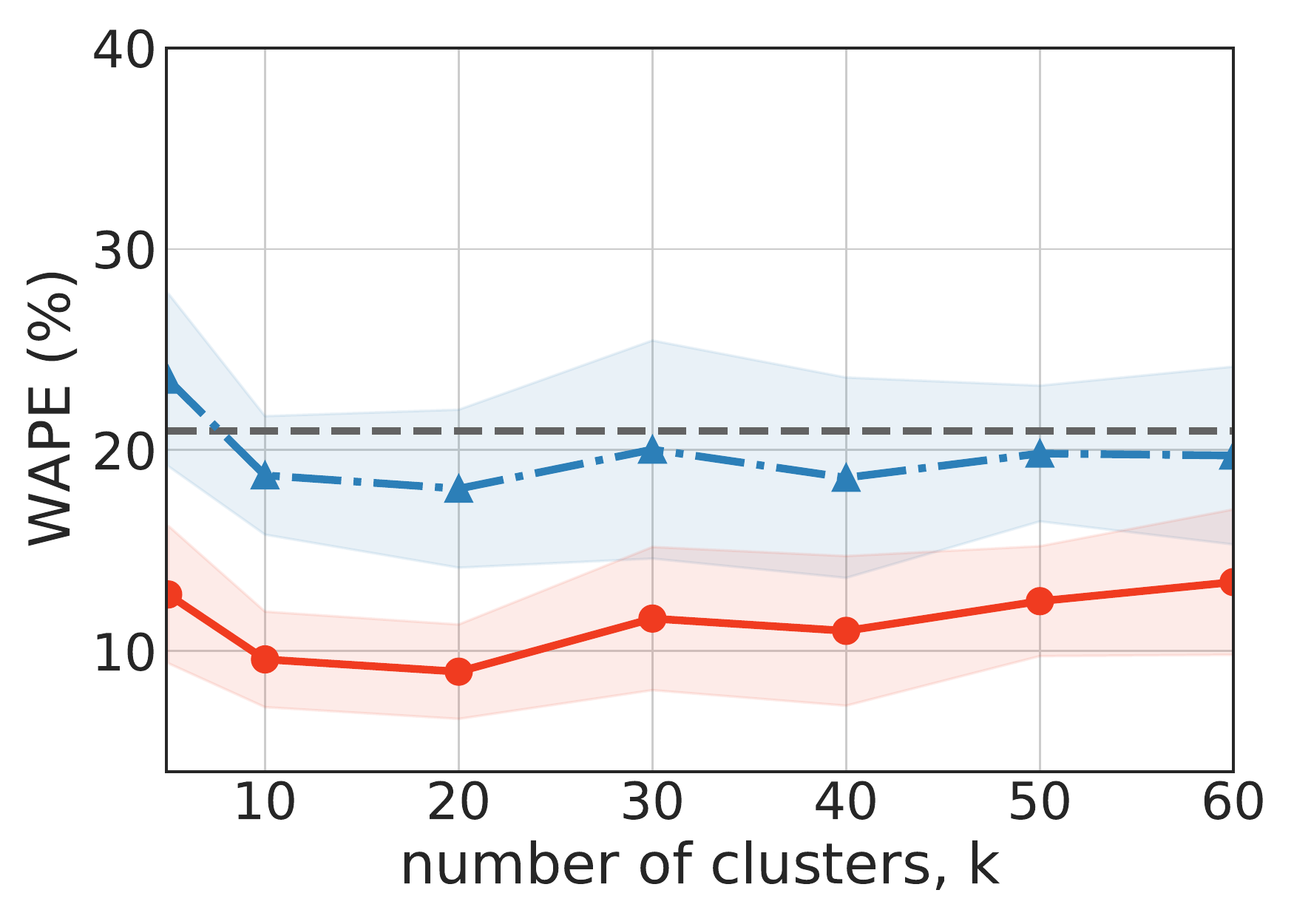} 	
		\end{overpic}
		%\caption{}
		%\label{fig:k20}
	\end{subfigure} \hspace{-0.6cm}
	~
	\begin{subfigure}[t]{0.33\textwidth}
	\centering
	\begin{overpic}[width=1\textwidth]{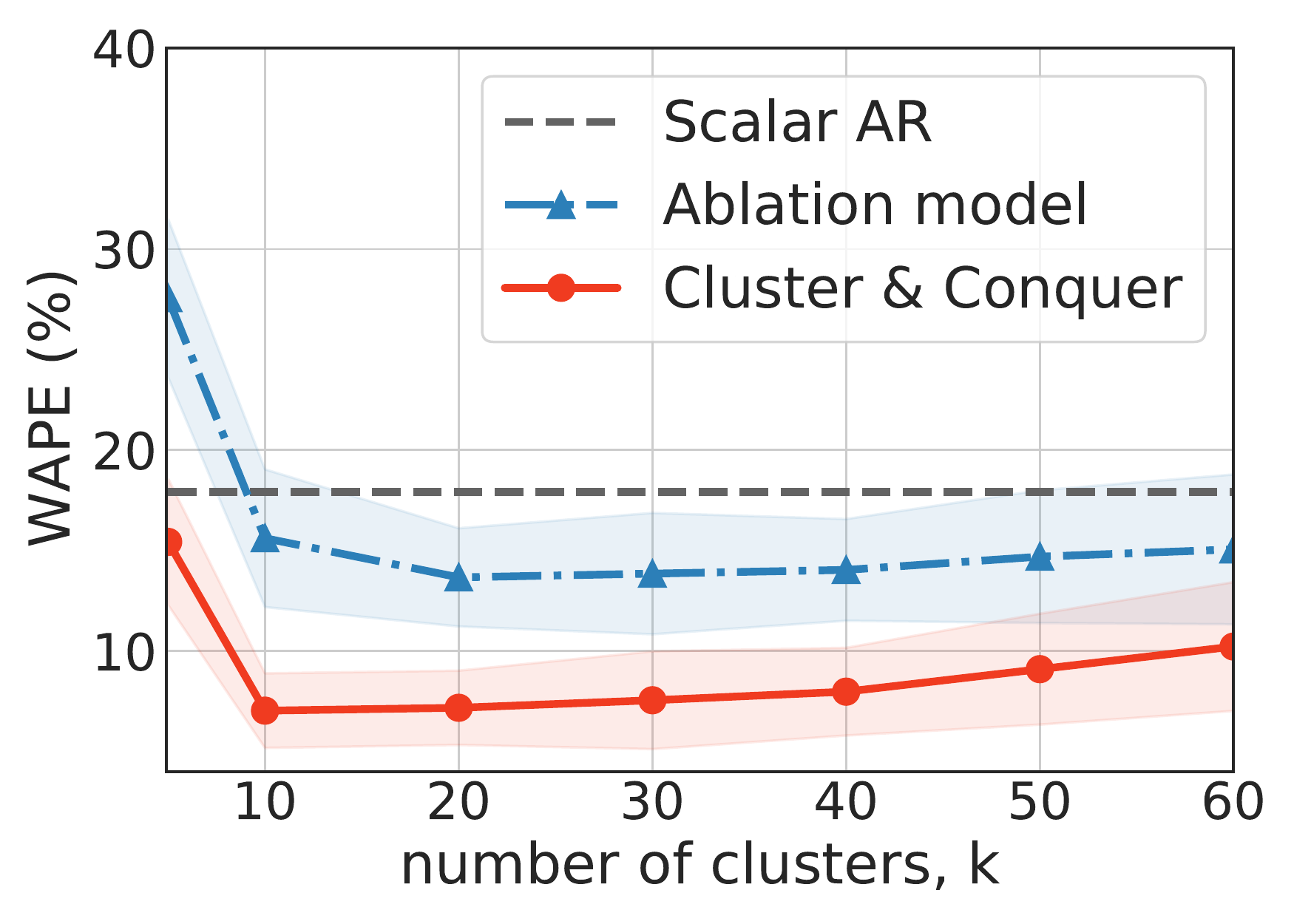} 	
	\end{overpic}
	%\caption{}
	%\label{fig:k30}
	\end{subfigure}

	\begin{subfigure}[t]{0.33\textwidth}
	\centering
	\begin{overpic}[width=1\textwidth]{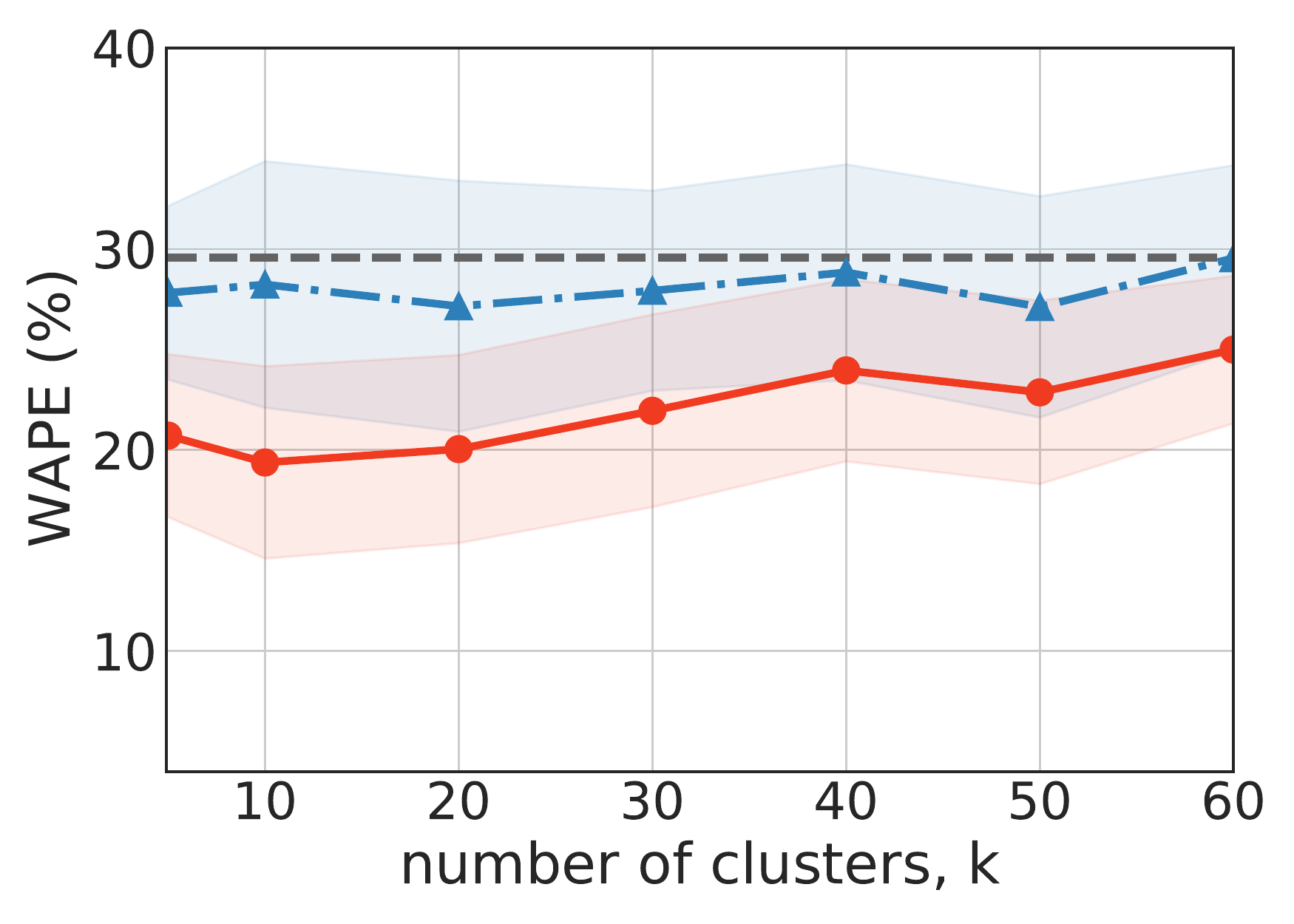}
	\end{overpic}
	%\caption{}
	%\label{fig:k10_p20}
\end{subfigure} \hspace{-0.6cm}
~
\begin{subfigure}[t]{0.33\textwidth}
	\centering
	\begin{overpic}[width=1\textwidth]{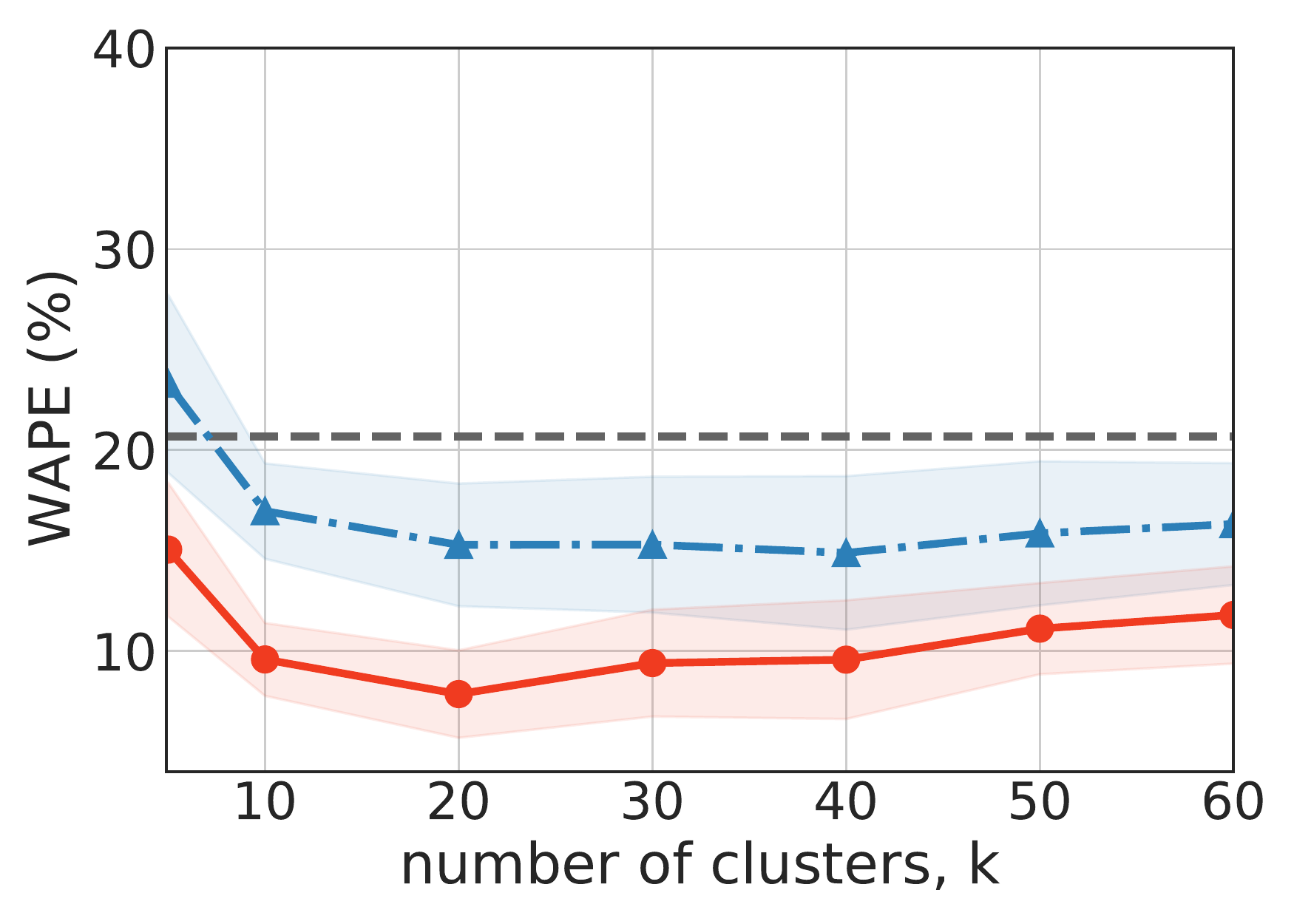} 	
	\end{overpic}
	%\caption{}
	%\label{fig:k20_p20}
\end{subfigure} \hspace{-0.6cm}
~
\begin{subfigure}[t]{0.33\textwidth}
	\centering
	\begin{overpic}[width=1\textwidth]{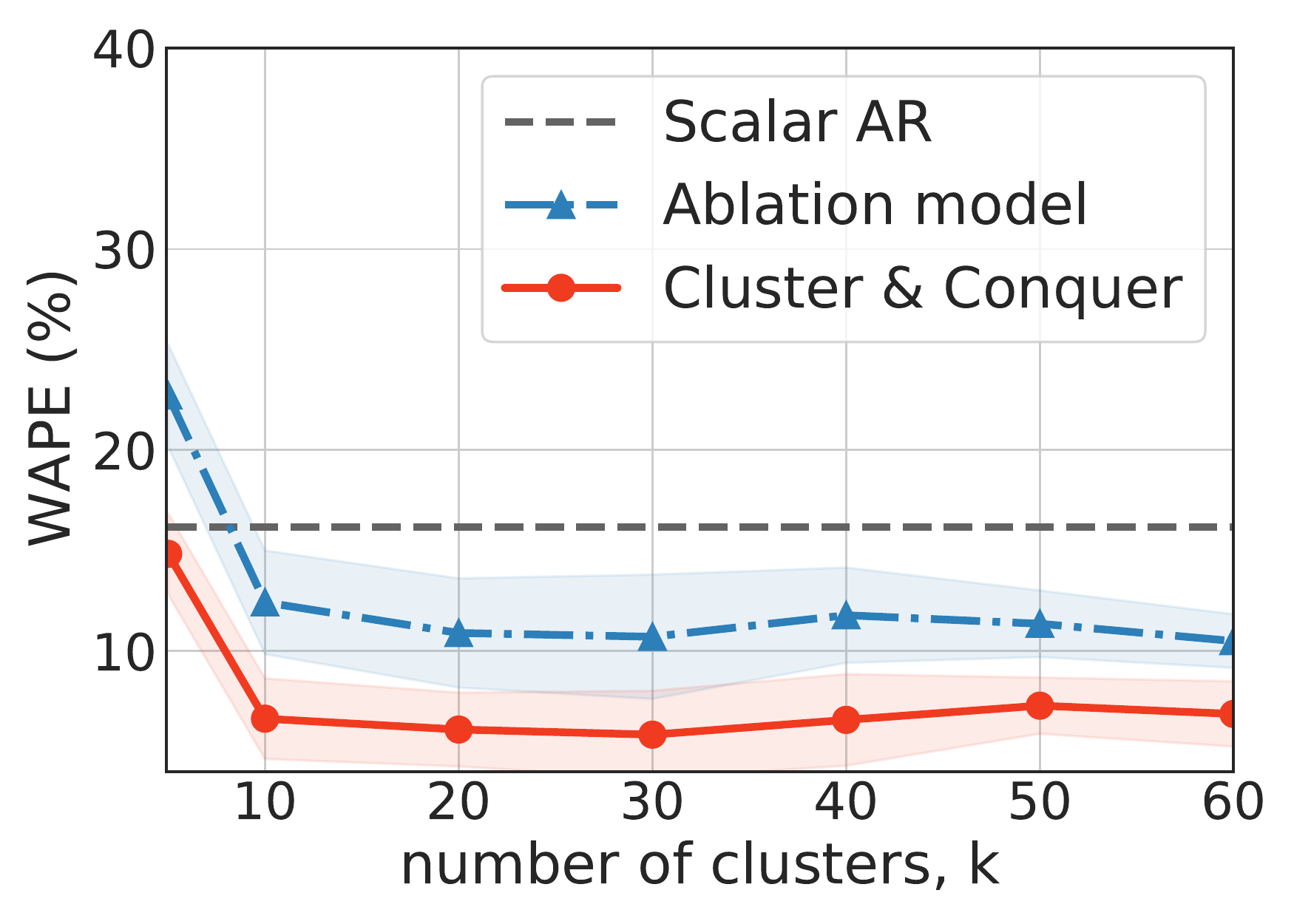} 	
	\end{overpic}
	%\caption{}
	%\label{fig:k30_p20}
\end{subfigure}	
	
	%\vspace{-0.3cm}
	
	\caption{\small Here we demonstrate the performance of \ours~on simulated data with different configurations. We evaluate our method as a function of the number of clusters and compare it to a Scalar AR model and an ablation model that uses random clusters. We see that \ours~is always better than Scalar AR and the ablation model.
	%
	%Further, we can notice that dips occur in locations where the number of clusters used by our algorithms correspond to actual number of clusters in the ground truth data.
	%
	%This shows that our algorithm faithfully recovers the correct clusters. 
	}
	\label{fig:simexp}
\end{figure*}

\begin{figure}[!b]
	\centering
	\begin{overpic}[width=0.43\textwidth]{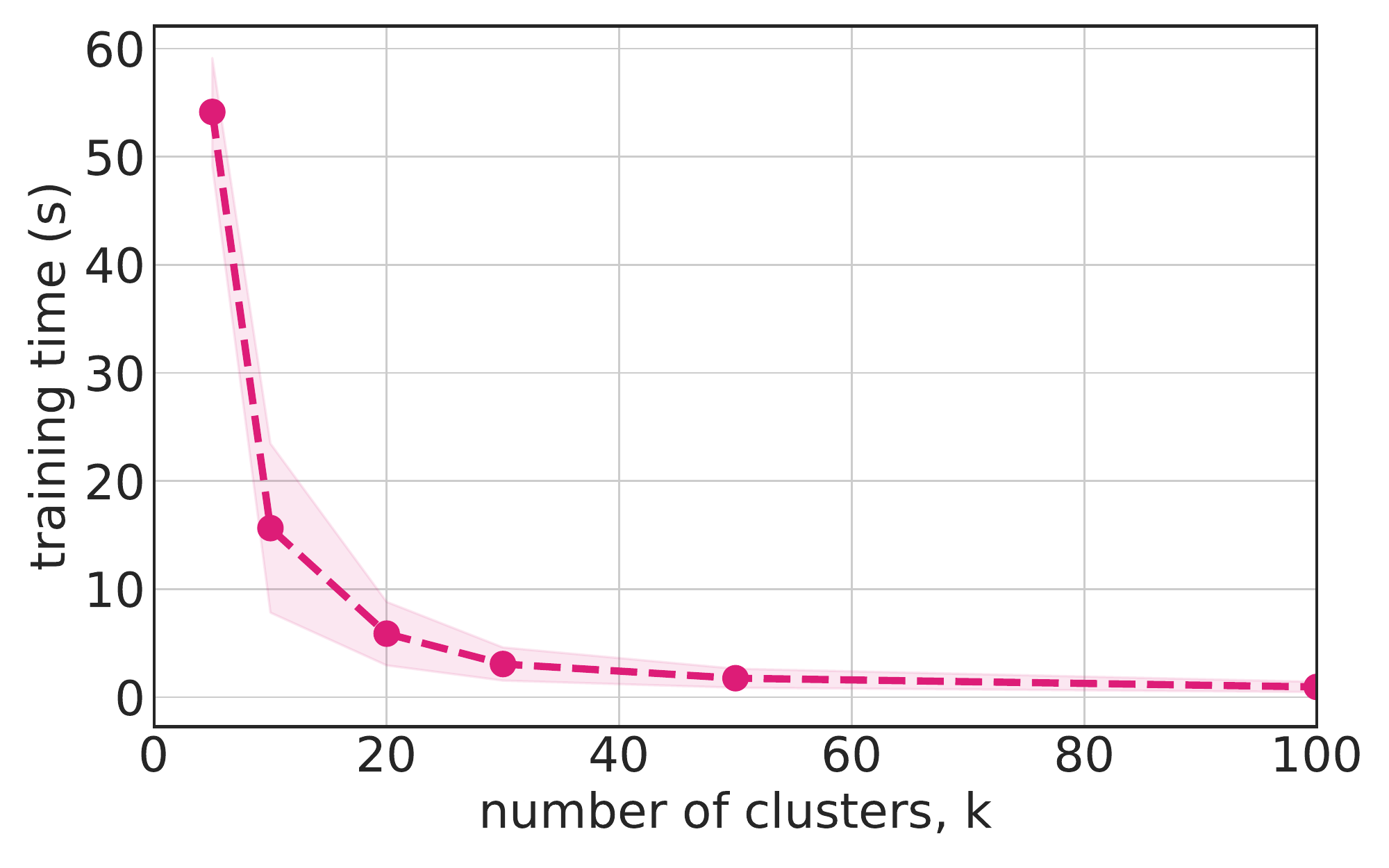}
	\end{overpic}
	\caption{Training time of VAR models on the whole dataset decreases with an increase in number of clusters. When a cluster is too big, VAR cannot scale as the computation complexity increases super-quadratically w.r.t number of time series in the VAR model.}
	\label{fig:timing}
\end{figure}

In this section we demonstrate the performance of our approach on both simulated and real-world datasets.
Specifically, we show that our framework, in combination with a simple linear function class (AR/VAR), is able to outperform several recently proposed deep learning methods for time-series forecasting. 

\paragraph{Setup.}
Given multivariate time series, we first use LinearSVR~\cite{fan2008liblinear} for training local AR models. 
Then we use the learned AR coefficients to cluster the time series. To do so, we  construct a $K$-nearest-neighbor graph ($K$=11) and then use the graph-clustering algorithm METIS~\cite{karypis1997metis} to form partitions. 
METIS is extremely scalable and has a sub-quadratic complexity in terms of the number of data points ($n$), unlike direct $k$-means or spectral clustering. 
Finally, we train global VAR models for each cluster using LinearSVR. 
We provide details about our implementation, including hyper-parameter settings, in the appendix in Section~\ref{sec:implement}. 
%Now we will present some representative results on a simulated setting. 
We refer to our algorithm as \ours. 

\paragraph{Experiments on simulated data.} 
We perform a simulation study to empirically verify that our algorithm works for non-i.i.d time series data given our generative assumptions. 
To do so, we consider several different configurations, i.e., we vary the number of clusters and the lag and period of the ground truth time series data.
In each case we construct dataset with $n=200$ time series.
Details about the data generation process are provided in the appendix in Section~\ref{sec:implement}.

Figure~\ref{fig:simexp} shows results for the different configurations, along with 1-standard deviation bands over 20 runs. 
The top row shows results for data with period and lag of 20, whereas the bottom row shows results for data that have period and lag of 10. 
The three columns correspond to ground truth datasets that have 10, 20 and 30 clusters, respectively, from left to right. 

To evaluate our model, we consider a rolling validation task with forecast horizon $24$ and with $7$ rolling windows. 
In each of the subplots, we plot the performance as measured by WAPE (see \eqref{eq:wape} in the appendix) as a function of the number of clusters ($k$) in our algorithm. 
We see that the performance of the algorithm is optimal or near-optimal at the correct value of $k$ (i.e., when $k$ is equal to number of clusters in the ground truth data). 
We also show that our method performs better compared to a Scalar AR model. Moreover, the performance is also better compared to an ablation model that considers random clusters followed by VAR within the clusters. 
We also note that the algorithm is robust to $k$ provided $k$ is not too far away from the number of clusters in the ground truth data.

In Figure~\ref{fig:timing}, we plot the running time of the algorithm as $k$ increases. As suspected when $k$ is small the running time is prohibitive as the VAR model in each cluster has to learn $O((n/k)^2)$ parameters. The shaded area corresponds to the standard deviation across 20 runs. 

\begin{table}[!ht]
\centering
\scalebox{0.85}{
\begin{tabular}{@{}lccc@{}}
\toprule
           & MAE & RMSE & MAPE  \\ \midrule        
FC-LSTM~\cite{sutskever2014sequence}  &   3.57                    & 6.2                      & 8.6   \\
SFM~\cite{zhang2017stock}     &    2.75                    & 4.32                     & 6.6   \\       
N-BEATS~\cite{oreshkin2019n}  &  3.41                    & 5.52                     & 7.65   \\         
DCRNN~\cite{li2017diffusion}   &  2.25                    & 4.04                     & 5.30   \\      
LSTNet~\cite{lai2018modeling}  &    2.34                    & 4.26                     & 5.41   \\     
ST-GCN~\cite{yu2017spatio}   &     2.25                    & 4.04                     & 5.26  \\      
TCN~\cite{bai2018empirical}        & 3.25                    & 5.51                     & 6.7   \\       
DeepState~\cite{rangapuram2018deep}  &  3.95                    & 6.49                     & 7.9  \\        
DeepGLO~\cite{sen2019think}    &  3.01                    & 5.25                     & 6.2      \\       
StemGNN~\cite{cao2020spectral}  &  2.14                    & 4.01                     & 5.01    \\
Scalar AR   &  2.13$\pm$0.080                    &  3.95$\pm$0.157                    &  4.88$\pm$0.2                  \\   
\ours    & \textbf{1.99}$\pm$0.030           & \textbf{3.68}$\pm$0.069            & \textbf{4.52}$\pm$0.1                      \\
 \bottomrule
\end{tabular}
}
\caption{\small Rolling validation results for the PEMS07 dataset. %\ours~outperforms state of the art deep models. For both \ours~and Scalar AR we repeat each experiment 10 times and report the one standard error.
} \label{tab:pems07}

\end{table}

\begin{table}[!ht]
\centering
\scalebox{0.85}{
\begin{tabular}{@{}lccc@{}}
\toprule
           & MAE & RMSE & MAPE \\ \midrule        
FC-LSTM~\cite{sutskever2014sequence}    &0.32                    & 0.54                     &31.0 \\
SFM~\cite{zhang2017stock}     & 0.17                    & 0.58                     &11.9   \\       
N-BEATS~\cite{oreshkin2019n}  & 0.08                    & 0.16                     &12.428                      \\         
LSTNet~\cite{lai2018modeling}     &  0.08                    & 0.12                     &12.74                     \\     
TCN~\cite{bai2018empirical}      &  0.1                     & 0.3                      &19.03                \\       
DeepState~\cite{rangapuram2018deep}    &  0.09                    & 0.76                     &19.21                      \\        
GraphWaveNet~\cite{wu2019graph}    & 0.19                    & 0.86                     &19.67                     \\  
DeepGLO~\cite{sen2019think}    &  0.09                    & 0.15                     &12.45                      \\       
StemGNN~\cite{cao2020spectral}  &  0.05                    & 0.07                     &10.58                      \\
Scalar AR  & 0.024$\pm$0.000                    & 0.041$\pm$0.000                    &5.67$\pm$0.1                     \\   
\ours   & \textbf{ 0.023}$\pm$0.000         & \textbf{0.040}$\pm$0.001             &\textbf{5.39}$\pm$0.0                        \\
 \bottomrule
\end{tabular}
}
\caption{\small Rolling validation results for the ECG dataset.} \label{tab:ecg}
\end{table}

\begin{table}[!ht]
\center
\scalebox{0.85}{
\begin{tabular}{@{}lcccccc@{}}
\toprule
           & WAPE & MAPE & SMAPE \\ \midrule        
DeepGLO~\cite{sen2019think}              & 8.2             & 34.1                      & 12.1                      \\
TCN (LeveledInit)~\cite{sen2019think}    & 9.2                    & 23.7                      & 12.6                          \\
LSTM                 & 10.9                    & 26.4                     & 15.4                  \\
DeepAR~\cite{salinas2019deepar}               &  8.6                    & 25.9                     & 14.1              \\
TCN~\cite{borovykh2017conditional}                  & 14.7                    & 47.6                     & 15.6                        \\
Prophet~\cite{prophet}              & 19.7                    & 39.3                     & 22.1                                \\
TRMF~\cite{yu2016temporal}                 & 10.4                    & 28.0                     & 15.1                           \\
SVD+TCN~\cite{sen2019think}              & 21.9                    & 43.7                     & 23.8                               \\
Scalar AR            & 8.11$\pm$0.14                    & 24.7$\pm$0.65                     & 11.2$\pm$0.19                                       \\
\ours                & \textbf{7.97}$\pm$0.09           & \textbf{20.48}$\pm$0.58            & \textbf{10.8}$\pm$0.08   \\
\bottomrule
\end{tabular}
}
\caption{\small Rolling val. results for the Electricity dataset.} \label{tab:electricity}
\end{table}

\begin{table}[!ht]
\centering
\scalebox{0.85}{
\begin{tabular}{@{}lcccccc@{}}
\toprule
           & WAPE & MAPE & SMAPE \\ \midrule        
DeepGLO~\cite{sen2019think}              &14.8                    & \textbf{16.8}                     & 14.2                      \\
TCN (LeveledInit)~\cite{sen2019think}                    &15.7                    & 20.1                     & 15.6                      \\
LSTM                     & 27.0                    & 35.7                     & 26.3                      \\
DeepAR~\cite{salinas2019deepar}                    & \textbf{14.0}                    & 20.1                     & \textbf{11.4}             \\
TCN~\cite{borovykh2017conditional}   & 20.4                    & 28.4                     & 23.6                     \\
Prophet~\cite{prophet}            & 30.3                    & 55.9                     & 42.0                      \\
TRMF~\cite{yu2016temporal}         & 15.9                    & 22.6                     & 18.1                     \\
SVD+TCN~\cite{sen2019think}        & 32.9                    & 68.7                     & 34.0                      \\
Scalar AR           &15.9$\pm$0.31                   & 20.0$\pm$1.14                     & 15.72$\pm$0.74                     \\
\ours    & \textbf{ 14.0}$\pm$0.16                    & \textbf{17.18}$\pm$0.49                    & 13.68$\pm$0.33   \\
\bottomrule
\end{tabular}
}
\caption{\small Rolling val. results for the Traffic dataset.} \label{tab:traffic}
\end{table}

\paragraph{\bf Experiments on real data.}
Here, we follow the setups described by~\citet{cao2020spectral} and~\citet{sen2019think}. 
We refer the reader to the appendix, particularly Section~\ref{sec:implement}, for details about the baselines and our implementation, as well as statistics of the datasets. 
For all datasets, we choose the number of clusters in \ours~to be $\floor{n/ 10}$, where $n$ is the number of time series. The number of clusters was manually tuned on the validation set between $\{\floor{n/ 5}, \floor{n/ 10}, \floor{n/ 20}\}$ and in all cases we found $\floor{n/ 10}$ to be the best.

First we compare the \ours~approach to baselines reported by~\cite{cao2020spectral}. We show results for the PEMS07~\cite{pemsbay} dataset in Table~\ref{tab:pems07}, and the ECG dataset~\cite{dau2019ucr} in Table~\ref{tab:ecg}.
The \ours~ approach outperforms all other models on all metrics, including state-of-the-art deep learning models. This is remarkable, because we only instantiate our framework with linear models.

Next, we compare \ours~against baselines reported by~\cite{sen2019think} for the Electricity and Traffic datasets, using the same metrics used in \cite{sen2019think} for a fair comparison. 
Table~\ref{tab:electricity} shows results for the Electricity dataset. Here, the advantage of the \ours~approach is pronounced, i.e., the gap between our approach and other deep learning models is distinct on all three metrics. 
On the other hand, the performance gap is less distinct for the Traffic dataset in Table~\ref{tab:traffic}.

Given these results, a few remarks are in place. First, it should be noted that the \ours~approach always improves over the scalar AR model. This shows that capturing intra-cluster dependencies is important for forecasting. Second, the \ours~approach is competitive to global deep learning models (e.g. DeepGLO, DCRNN, LSTNet). One reason for that is that the \ours~approach suppresses noise from irrelevant time series when predicting the future of a particular time series. Third, the \ours~approach does not use specially designed covariates. For instance, DeepAR~\cite{salinas2019deepar} uses specially designed time-based covariates such as transformations of days of the week, and holidays.

\section{Conclusions}
\label{sec:conclusion}
We proposed a three-stage framework, based on a divide and conquer principle, for large-scale multivariate time-series forecasting. Our \ours~framework is highly flexible, i.e., it allows for the use of any multivariate time-series model coupled with a scalable clustering algorithm. 
In turn, this framework is computationally efficient, embarrassingly parallel, and typically more efficient than dense vector auto regression.
We provide theoretical guarantees for our framework in a mixed linear regression setting. Experiments on both synthetic and multiple real-world datasets show that the \ours~ approach achieves state-of-the-art results compared to baselines.

\bibliography{references}

\begin{thebibliography}{45}
\providecommand{\natexlab}[1]{#1}
\providecommand{\url}[1]{\texttt{#1}}
\expandafter\ifx\csname urlstyle\endcsname\relax
  \providecommand{\doi}[1]{doi: #1}\else
  \providecommand{\doi}{doi: \begingroup \urlstyle{rm}\Url}\fi

\bibitem[Aghabozorgi et~al.(2015)Aghabozorgi, Seyed~Shirkhorshidi, and
  Ying~Wah]{aghabozorgi2015time}
S.~Aghabozorgi, A.~Seyed~Shirkhorshidi, and T.~Ying~Wah.
\newblock Time-series clustering - a decade review.
\newblock \emph{Inf. Syst.}, 53\penalty0 (C):\penalty0 16–38, Oct. 2015.
\newblock ISSN 0306-4379.
\newblock \doi{10.1016/j.is.2015.04.007}.
\newblock URL \url{https://doi.org/10.1016/j.is.2015.04.007}.

\bibitem[Azencot et~al.(2020)Azencot, Erichson, Lin, and
  Mahoney]{azencot2020forecasting}
O.~Azencot, N.~B. Erichson, V.~Lin, and M.~Mahoney.
\newblock Forecasting sequential data using consistent koopman autoencoders.
\newblock In \emph{International Conference on Machine Learning}, pages
  475--485. PMLR, 2020.

\bibitem[Bai et~al.(2020)Bai, Yao, Li, Wang, and Wang]{bai2020adaptive}
L.~Bai, L.~Yao, C.~Li, X.~Wang, and C.~Wang.
\newblock Adaptive graph convolutional recurrent network for traffic
  forecasting.
\newblock \emph{Advances in Neural Information Processing Systems}, 33, 2020.

\bibitem[Bai et~al.(2018)Bai, Kolter, and Koltun]{bai2018empirical}
S.~Bai, J.~Z. Kolter, and V.~Koltun.
\newblock An empirical evaluation of generic convolutional and recurrent
  networks for sequence modeling.
\newblock \emph{CoRR}, abs/1803.01271, 2018.
\newblock URL \url{http://arxiv.org/abs/1803.01271}.

\bibitem[Bandara et~al.(2020)Bandara, Bergmeir, and
  Smyl]{bandara2020forecasting}
K.~Bandara, C.~Bergmeir, and S.~Smyl.
\newblock Forecasting across time series databases using recurrent neural
  networks on groups of similar series: {A} clustering approach.
\newblock \emph{Expert Syst. Appl.}, 140, 2020.
\newblock \doi{10.1016/j.eswa.2019.112896}.
\newblock URL \url{https://doi.org/10.1016/j.eswa.2019.112896}.

\bibitem[Benidis et~al.(2020)Benidis, Rangapuram, Flunkert, Wang, Maddix,
  Turkmen, Gasthaus, Bohlke-Schneider, Salinas, Stella,
  et~al.]{benidis2020neural}
K.~Benidis, S.~S. Rangapuram, V.~Flunkert, B.~Wang, D.~Maddix, C.~Turkmen,
  J.~Gasthaus, M.~Bohlke-Schneider, D.~Salinas, L.~Stella, et~al.
\newblock Neural forecasting: Introduction and literature overview.
\newblock \emph{arXiv preprint arXiv:2004.10240}, 2020.

\bibitem[Bhatia et~al.(2010)]{bhatia2010survey}
N.~Bhatia et~al.
\newblock Survey of nearest neighbor techniques.
\newblock \emph{arXiv preprint arXiv:1007.0085}, 2010.

\bibitem[Borovykh et~al.(2017)Borovykh, Bohte, and
  Oosterlee]{borovykh2017conditional}
A.~Borovykh, S.~Bohte, and C.~W. Oosterlee.
\newblock Conditional time series forecasting with convolutional neural
  networks.
\newblock \emph{arXiv preprint arXiv:1703.04691}, 2017.

\bibitem[Cao et~al.(2020)Cao, Wang, Duan, Zhang, Zhu, Huang, Tong, Xu, Bai,
  Tong, and Zhang]{cao2020spectral}
D.~Cao, Y.~Wang, J.~Duan, C.~Zhang, X.~Zhu, C.~Huang, Y.~Tong, B.~Xu, J.~Bai,
  J.~Tong, and Q.~Zhang.
\newblock Spectral temporal graph neural network for multivariate time-series
  forecasting.
\newblock In H.~Larochelle, M.~Ranzato, R.~Hadsell, M.~Balcan, and H.~Lin,
  editors, \emph{Advances in Neural Information Processing Systems 33: Annual
  Conference on Neural Information Processing Systems 2020, NeurIPS 2020,
  December 6-12, 2020, virtual}, 2020.
\newblock URL
  \url{https://proceedings.neurips.cc/paper/2020/hash/cdf6581cb7aca4b7e19ef136c6e601a5-Abstract.html}.

\bibitem[Chen et~al.(2001)Chen, Petty, Skabardonis, Varaiya, and Jia]{pemsbay}
C.~Chen, K.~Petty, A.~Skabardonis, P.~Varaiya, and Z.~Jia.
\newblock Freeway performance measurement system: Mining loop detector data.
\newblock \emph{Transportation Research Record}, 1748\penalty0 (1):\penalty0
  96--102, 2001.
\newblock \doi{10.3141/1748-12}.
\newblock URL \url{https://doi.org/10.3141/1748-12}.

\bibitem[Dau et~al.(2019)Dau, Bagnall, Kamgar, Yeh, Zhu, Gharghabi,
  Ratanamahatana, and Keogh]{dau2019ucr}
H.~A. Dau, A.~J. Bagnall, K.~Kamgar, C.~M. Yeh, Y.~Zhu, S.~Gharghabi, C.~A.
  Ratanamahatana, and E.~J. Keogh.
\newblock The {UCR} time series archive.
\newblock \emph{{IEEE} {CAA} J. Autom. Sinica}, 6\penalty0 (6):\penalty0
  1293--1305, 2019.
\newblock \doi{10.1109/jas.2019.1911747}.
\newblock URL \url{https://doi.org/10.1109/jas.2019.1911747}.

\bibitem[Ding et~al.(2015)Ding, Wang, Dang, Fu, Zhang, and
  Zhang]{ding2015yading}
R.~Ding, Q.~Wang, Y.~Dang, Q.~Fu, H.~Zhang, and D.~Zhang.
\newblock Yading: fast clustering of large-scale time series data.
\newblock \emph{Proceedings of the VLDB Endowment}, 8\penalty0 (5):\penalty0
  473--484, 2015.

\bibitem[Divine(2012)]{divine2012climate}
D.~V. Divine.
\newblock M. mudelsee, , climate time series analysis: Classical statistical
  and bootstrap methods {(2010)} springer, dordrecht 978-90-481-9482-7.
\newblock \emph{Comput. Geosci.}, 43:\penalty0 24, 2012.
\newblock \doi{10.1016/j.cageo.2012.02.030}.
\newblock URL \url{https://doi.org/10.1016/j.cageo.2012.02.030}.

\bibitem[Erichson et~al.(2019)Erichson, Muehlebach, and
  Mahoney]{erichson2019physics}
N.~B. Erichson, M.~Muehlebach, and M.~W. Mahoney.
\newblock Physics-informed autoencoders for lyapunov-stable fluid flow
  prediction.
\newblock \emph{arXiv preprint arXiv:1905.10866}, 2019.

\bibitem[Erichson et~al.(2021)Erichson, Azencot, Queiruga, Hodgkinson, and
  Mahoney]{lipschitzrnn}
N.~B. Erichson, O.~Azencot, A.~Queiruga, L.~Hodgkinson, and M.~W. Mahoney.
\newblock Lipschitz recurrent neural networks.
\newblock In \emph{International Conference on Learning Representations}, 2021.
\newblock URL \url{https://openreview.net/forum?id=-N7PBXqOUJZ}.

\bibitem[Facebook(2020)]{prophet}
Facebook.
\newblock Fb prophet, forecasting at scale.
\newblock \url{https://github.com/facebook/prophet}, 2020.

\bibitem[Faloutsos et~al.(2020)Faloutsos, Flunkert, Gasthaus, Januschowski, and
  Wang]{faloutsos2020forecasting}
C.~Faloutsos, V.~Flunkert, J.~Gasthaus, T.~Januschowski, and Y.~Wang.
\newblock Forecasting big time series: Theory and practice.
\newblock In A.~E.~F. Seghrouchni, G.~Sukthankar, T.~Liu, and M.~van Steen,
  editors, \emph{Companion of The 2020 Web Conference 2020, Taipei, Taiwan,
  April 20-24, 2020}, pages 320--321. {ACM} / {IW3C2}, 2020.
\newblock \doi{10.1145/3366424.3383118}.
\newblock URL \url{https://doi.org/10.1145/3366424.3383118}.

\bibitem[Fan et~al.(2008)Fan, Chang, Hsieh, Wang, and Lin]{fan2008liblinear}
R.~Fan, K.~Chang, C.~Hsieh, X.~Wang, and C.~Lin.
\newblock {LIBLINEAR:} {A} library for large linear classification.
\newblock \emph{J. Mach. Learn. Res.}, 9:\penalty0 1871--1874, 2008.
\newblock URL \url{https://dl.acm.org/citation.cfm?id=1442794}.

\bibitem[Goldenshluger and Zeevi(2001)]{goldenshluger2001nonasymptotic}
A.~Goldenshluger and A.~Zeevi.
\newblock Nonasymptotic bounds for autoregressive time series modeling.
\newblock \emph{Ann. Statist.}, 29\penalty0 (2):\penalty0 417--444, 2001.
\newblock ISSN 0090-5364.
\newblock \doi{10.1214/aos/1009210547}.
\newblock URL \url{https://doi.org/10.1214/aos/1009210547}.

\bibitem[Guo et~al.(2016)Guo, Wang, and Yao]{guo2016high}
S.~Guo, Y.~Wang, and Q.~Yao.
\newblock High-dimensional and banded vector autoregressions.
\newblock \emph{Biometrika}, 103\penalty0 (4):\penalty0 889--903, 2016.
\newblock ISSN 0006-3444.
\newblock \doi{10.1093/biomet/asw046}.
\newblock URL \url{https://doi.org/10.1093/biomet/asw046}.

\bibitem[Hyndman and Athanasopoulos(2018)]{hyndman2018forecasting}
R.~J. Hyndman and G.~Athanasopoulos.
\newblock \emph{Forecasting: principles and practice}.
\newblock OTexts, 2018.

\bibitem[Kalpakis et~al.(2001)Kalpakis, Gada, and
  Puttagunta]{kalpakis2001distance}
K.~Kalpakis, D.~Gada, and V.~Puttagunta.
\newblock Distance measures for effective clustering of {ARIMA} time-series.
\newblock In N.~Cercone, T.~Y. Lin, and X.~Wu, editors, \emph{Proceedings of
  the 2001 {IEEE} International Conference on Data Mining, 29 November - 2
  December 2001, San Jose, California, {USA}}, pages 273--280. {IEEE} Computer
  Society, 2001.
\newblock \doi{10.1109/ICDM.2001.989529}.
\newblock URL \url{https://doi.org/10.1109/ICDM.2001.989529}.

\bibitem[Kannan and Vempala(2008)]{kannan2009spectral}
R.~Kannan and S.~Vempala.
\newblock Spectral algorithms.
\newblock \emph{Found. Trends Theor. Comput. Sci.}, 4\penalty0 (3-4):\penalty0
  front matter, 157--288 (2009), 2008.
\newblock ISSN 1551-305X.
\newblock \doi{10.1561/0400000025}.
\newblock URL \url{https://doi.org/10.1561/0400000025}.

\bibitem[Karypis and Kumar(1998)]{karypis1997metis}
G.~Karypis and V.~Kumar.
\newblock \emph{{MeTiS}: A Software Package for Partitioning Unstructured
  Graphs, Partitioning Meshes, and Computing Fill-Reducing Orderings of Sparse
  Matrices}.
\newblock Department of Computer Science, University of Minnesota, version 4.0
  edition, Sept. 1998.

\bibitem[Kuznetsov and Mohri(2015)]{kuznetsov2015learning}
V.~Kuznetsov and M.~Mohri.
\newblock Learning theory and algorithms for forecasting non-stationary time
  series.
\newblock In \emph{NIPS}, pages 541--549. Citeseer, 2015.

\bibitem[Lai et~al.(2018)Lai, Chang, Yang, and Liu]{lai2018modeling}
G.~Lai, W.~Chang, Y.~Yang, and H.~Liu.
\newblock Modeling long- and short-term temporal patterns with deep neural
  networks.
\newblock In K.~Collins{-}Thompson, Q.~Mei, B.~D. Davison, Y.~Liu, and
  E.~Yilmaz, editors, \emph{The 41st International {ACM} {SIGIR} Conference on
  Research {\&} Development in Information Retrieval, {SIGIR} 2018, Ann Arbor,
  MI, USA, July 08-12, 2018}, pages 95--104. {ACM}, 2018.
\newblock \doi{10.1145/3209978.3210006}.
\newblock URL \url{https://doi.org/10.1145/3209978.3210006}.

\bibitem[Li et~al.(2017)Li, Yu, Shahabi, and Liu]{li2017diffusion}
Y.~Li, R.~Yu, C.~Shahabi, and Y.~Liu.
\newblock Diffusion convolutional recurrent neural network: Data-driven traffic
  forecasting.
\newblock \emph{arXiv preprint arXiv:1707.01926}, 2017.

\bibitem[Li et~al.(2018)Li, Yu, Shahabi, and Liu]{li2018diffusion}
Y.~Li, R.~Yu, C.~Shahabi, and Y.~Liu.
\newblock Diffusion convolutional recurrent neural network: Data-driven traffic
  forecasting.
\newblock In \emph{International Conference on Learning Representations}, 2018.

\bibitem[Lim et~al.(2021)Lim, Erichson, Hodgkinson, and Mahoney]{lim2021noisy}
S.~H. Lim, N.~B. Erichson, L.~Hodgkinson, and M.~W. Mahoney.
\newblock Noisy recurrent neural networks.
\newblock \emph{arXiv preprint arXiv:2102.04877}, 2021.

\bibitem[L{\"{o}}ffler et~al.(2019)L{\"{o}}ffler, Zhang, and
  Zhou]{loffler2019optimality}
M.~L{\"{o}}ffler, A.~Y. Zhang, and H.~H. Zhou.
\newblock Optimality of spectral clustering for gaussian mixture model.
\newblock \emph{CoRR}, abs/1911.00538, 2019.
\newblock URL \url{http://arxiv.org/abs/1911.00538}.

\bibitem[Matsubara et~al.(2014)Matsubara, Sakurai, van Panhuis, and
  Faloutsos]{matsubara2014funnel}
Y.~Matsubara, Y.~Sakurai, W.~G. van Panhuis, and C.~Faloutsos.
\newblock {FUNNEL:} automatic mining of spatially coevolving epidemics.
\newblock In S.~A. Macskassy, C.~Perlich, J.~Leskovec, W.~Wang, and R.~Ghani,
  editors, \emph{The 20th {ACM} {SIGKDD} International Conference on Knowledge
  Discovery and Data Mining, {KDD} '14, New York, NY, {USA} - August 24 - 27,
  2014}, pages 105--114. {ACM}, 2014.
\newblock \doi{10.1145/2623330.2623624}.
\newblock URL \url{https://doi.org/10.1145/2623330.2623624}.

\bibitem[McDonald et~al.(2017)McDonald, Shalizi, and
  Schervish]{mcdonald2017nonparametric}
D.~J. McDonald, C.~R. Shalizi, and M.~Schervish.
\newblock Nonparametric risk bounds for time-series forecasting.
\newblock \emph{J. Mach. Learn. Res.}, 18:\penalty0 Paper No. 32, 40, 2017.
\newblock ISSN 1532-4435.

\bibitem[Oreshkin et~al.(2020)Oreshkin, Carpov, Chapados, and
  Bengio]{oreshkin2019n}
B.~N. Oreshkin, D.~Carpov, N.~Chapados, and Y.~Bengio.
\newblock {N-BEATS:} neural basis expansion analysis for interpretable time
  series forecasting.
\newblock In \emph{8th International Conference on Learning Representations,
  {ICLR} 2020, Addis Ababa, Ethiopia, April 26-30, 2020}. OpenReview.net, 2020.
\newblock URL \url{https://openreview.net/forum?id=r1ecqn4YwB}.

\bibitem[Rangapuram et~al.(2018)Rangapuram, Seeger, Gasthaus, Stella, Wang, and
  Januschowski]{rangapuram2018deep}
S.~S. Rangapuram, M.~W. Seeger, J.~Gasthaus, L.~Stella, Y.~Wang, and
  T.~Januschowski.
\newblock Deep state space models for time series forecasting.
\newblock In S.~Bengio, H.~M. Wallach, H.~Larochelle, K.~Grauman,
  N.~Cesa{-}Bianchi, and R.~Garnett, editors, \emph{Advances in Neural
  Information Processing Systems 31: Annual Conference on Neural Information
  Processing Systems 2018, NeurIPS 2018, December 3-8, 2018, Montr{\'{e}}al,
  Canada}, pages 7796--7805, 2018.
\newblock URL
  \url{https://proceedings.neurips.cc/paper/2018/hash/5cf68969fb67aa6082363a6d4e6468e2-Abstract.html}.

\bibitem[Salinas et~al.(2019)Salinas, Flunkert, Gasthaus, and
  Januschowski]{salinas2019deepar}
D.~Salinas, V.~Flunkert, J.~Gasthaus, and T.~Januschowski.
\newblock Deepar: Probabilistic forecasting with autoregressive recurrent
  networks.
\newblock \emph{International Journal of Forecasting}, 2019.

\bibitem[Sen et~al.(2019)Sen, Yu, and Dhillon]{sen2019think}
R.~Sen, H.~Yu, and I.~S. Dhillon.
\newblock Think globally, act locally: {A} deep neural network approach to
  high-dimensional time series forecasting.
\newblock In H.~M. Wallach, H.~Larochelle, A.~Beygelzimer,
  F.~d'Alch{\'{e}}{-}Buc, E.~B. Fox, and R.~Garnett, editors, \emph{Advances in
  Neural Information Processing Systems 32: Annual Conference on Neural
  Information Processing Systems 2019, NeurIPS 2019, December 8-14, 2019,
  Vancouver, BC, Canada}, pages 4838--4847, 2019.
\newblock URL
  \url{https://proceedings.neurips.cc/paper/2019/hash/3a0844cee4fcf57de0c71e9ad3035478-Abstract.html}.

\bibitem[Sutskever et~al.(2014)Sutskever, Vinyals, and
  Le]{sutskever2014sequence}
I.~Sutskever, O.~Vinyals, and Q.~V. Le.
\newblock Sequence to sequence learning with neural networks.
\newblock In Z.~Ghahramani, M.~Welling, C.~Cortes, N.~D. Lawrence, and K.~Q.
  Weinberger, editors, \emph{Advances in Neural Information Processing Systems
  27: Annual Conference on Neural Information Processing Systems 2014, December
  8-13 2014, Montreal, Quebec, Canada}, pages 3104--3112, 2014.
\newblock URL
  \url{https://proceedings.neurips.cc/paper/2014/hash/a14ac55a4f27472c5d894ec1c3c743d2-Abstract.html}.

\bibitem[Venkataramana~Kini and Chandra~Sekhar(2013)]{kini2013bayesian}
B.~Venkataramana~Kini and C.~Chandra~Sekhar.
\newblock Bayesian mixture of {AR} models for time series clustering.
\newblock \emph{PAA Pattern Anal. Appl.}, 16\penalty0 (2):\penalty0 179--200,
  2013.
\newblock ISSN 1433-7541.
\newblock \doi{10.1007/s10044-011-0247-5}.
\newblock URL \url{https://doi.org/10.1007/s10044-011-0247-5}.

\bibitem[Warren~Liao(2005)]{liao2005clustering}
T.~Warren~Liao.
\newblock Clustering of time series data-a survey.
\newblock \emph{Pattern Recogn.}, 38\penalty0 (11):\penalty0 1857–1874, Nov.
  2005.
\newblock ISSN 0031-3203.
\newblock \doi{10.1016/j.patcog.2005.01.025}.
\newblock URL \url{https://doi.org/10.1016/j.patcog.2005.01.025}.

\bibitem[Wu et~al.(2019)Wu, Pan, Long, Jiang, and Zhang]{wu2019graph}
Z.~Wu, S.~Pan, G.~Long, J.~Jiang, and C.~Zhang.
\newblock Graph wavenet for deep spatial-temporal graph modeling.
\newblock In \emph{IJCAI}, 2019.

\bibitem[Yu et~al.(2017)Yu, Yin, and Zhu]{yu2017spatio}
B.~Yu, H.~Yin, and Z.~Zhu.
\newblock Spatio-temporal graph convolutional networks: A deep learning
  framework for traffic forecasting.
\newblock \emph{arXiv preprint arXiv:1709.04875}, 2017.

\bibitem[Yu et~al.(2018)Yu, Yin, and Zhu]{yu2018spatio}
B.~Yu, H.~Yin, and Z.~Zhu.
\newblock Spatio-temporal graph convolutional networks: a deep learning
  framework for traffic forecasting.
\newblock In \emph{Proceedings of the 27th International Joint Conference on
  Artificial Intelligence}, pages 3634--3640, 2018.

\bibitem[Yu et~al.(2016)Yu, Rao, and Dhillon]{yu2016temporal}
H.-F. Yu, N.~Rao, and I.~S. Dhillon.
\newblock Temporal regularized matrix factorization for high-dimensional time
  series prediction.
\newblock In \emph{Advances in neural information processing systems}, pages
  847--855, 2016.

\bibitem[Zhang et~al.(2017)Zhang, Aggarwal, and Qi]{zhang2017stock}
L.~Zhang, C.~C. Aggarwal, and G.~Qi.
\newblock Stock price prediction via discovering multi-frequency trading
  patterns.
\newblock In \emph{Proceedings of the 23rd {ACM} {SIGKDD} International
  Conference on Knowledge Discovery and Data Mining, Halifax, NS, Canada,
  August 13 - 17, 2017}, pages 2141--2149. {ACM}, 2017.
\newblock \doi{10.1145/3097983.3098117}.
\newblock URL \url{https://doi.org/10.1145/3097983.3098117}.

\bibitem[Zhu and Shasha(2002)]{zhu2002statstream}
Y.~Zhu and D.~E. Shasha.
\newblock Statstream: Statistical monitoring of thousands of data streams in
  real time.
\newblock In \emph{Proceedings of 28th International Conference on Very Large
  Data Bases, {VLDB} 2002, Hong Kong, August 20-23, 2002}, pages 358--369.
  Morgan Kaufmann, 2002.
\newblock \doi{10.1016/B978-155860869-6/50039-1}.
\newblock URL \url{http://www.vldb.org/conf/2002/S10P04.pdf}.

\end{thebibliography}

\clearpage
\appendix

\section{Proof of result for near mixture of Gaussians}
\label{sec:thm-proof}

\paragraph{Notation:}
Let us define the following loss function, $\ell \colon [\numclus]^{[\numdata]} \times
[\numclus]^{[\numdata]} \to \R_+$, given by
\[
\ell(\mathsf{c}, \mathsf{c}') \defn
\min_{\varphi}\sum_{i=1}^\numdata \1\Big\{\mathsf{c}(i) \neq \varphi(\mathsf{c}'(i))\Big\},
\quad \mbox{for any}~ \mathsf{c}, \mathsf{c}' \in [\numclus]^{[\numdata]},
\]
where the minimum ranges over bijections $\varphi \colon [\numclus] \to [\numclus]$.
Intuitively, this measures the number disagreements between the cluster labellings
$\mathsf{c}, \mathsf{c}'$.
Let us set
\[
\Delta \defn \min_{i \neq j \in [k]} \|\optcenter_i - \optcenter_j\|_2
\]

\paragraph{Proof:}
Since Algorithms~\ref{alg:spectral-clustering}
and~\ref{alg:low-rank} differ only by the application of
the orthogonal matrix $\hat U$, it is
striaghtforward to see that
there is a bijection $\phi \colon [k] \to [k]$ such that
$\anclus{i} = \phi(\predclus{i})$ for each $i \in [n]$.
Therefore, by definition of
$\ell(\cdot, \cdot)$, we have that
\[
\ell(\hat{\mathsf{c}}_{\rm spec}, \mathsf{c}^\star)
= \ell(\hat{\mathsf{c}}_{\rm appx}, \mathsf{c}^\star).
\]
Suppose that the following good event holds:
\[
\concevent \defn \Big\{\opnorm{G} \leq \sqrt{2\eigmax}(\sqrt{\numdata} + \sqrt{\dimension})\Big\}.
\]
Consider the sets
\[
\badindices \defn \Big\{ i \in [\numdata]
: \|\optcenter_{\trueclus{i}} - \hat \theta_{i}\|_2 \geq \Delta/2
\Big\} \qquad \mbox{and} \qquad
\goodindices \defn [n] \setminus \badindices.
\]
Under $\concevent$, it is possible to show that
\[
|\badindices| \leq
\frac{128 k \opnorm{G}^2}{\Delta^2}
\leq
\frac{512 \eigmax k (\numdata + \dimension)}{\Delta^2}.
\]
(The first inequality above is Proposition~\ref{prop:bad-ind-card}.)
We can partition the good indices into disjoint sets $I_j$ as follows:
\[
I_j \defn \Big\{ i \in [\numdata] :
\trueclus{i} = j,~ \|\hat \theta_i - \optcenter_{\trueclus{i}} \|_2 < \Delta/2\Big\},
\qquad
j =1, \ldots, \numclus.
\]
Note that under the assumption on $\Delta$, $|\badindices| \leq \tfrac{\beta n}{2k}$,
which implies that $I_j$ are nonempty for all $j \in [\numclus]$.
Additionally, $\anclus{I_j}$ are pairwise disjoint, as we demonstrate
in Lemma~\ref{lem:disjointness}, after which it follows that
there exists a permutation $\varphi$ such that
\[
\varphi(\anclus{i}) = j, \quad \mbox{for each}~i\in I_j.
\]
This implies that under $\concevent$
\[
\ell(\hat{\mathsf{c}}_{\rm appx}, \mathsf{c}^\star)
\leq |\badindices| \leq
\frac{512 \eigmax k (\numdata + \dimension)}{\Delta^2} \leq \frac{1}{2},
\]
which implies that $\hat{\mathsf{c}}_{\rm appx}, \mathsf{c}^\star$ are
equal up to permutation. The result now follows by noting
that $\P(\concevent) \geq 1 - \e^{-0.08 n}$, as we show in
Lemma~\ref{lem:conc}.

\subsection{Auxillary results for Theorem~\ref{thm:main-GMM-result}}

We assume that for all $i$, we have $\Sigma_i \preceq \eigmax I_\dimension$ for
some constant $\eigmax > 0$. It will be convenient for us to consider the sample matrix
$X \in \R^{\dimension \times \numdata}$, which has columns $X_1, \dots, X_\numdata \in \R^d$.
Then it is easy to see that
\[
X = \E X + (X - \E X) \eqcolon \OptCenterMat + \NoiseMat,
\]
where $\OptCenterMat, \NoiseMat \in \R^{\dimension \times \numdata}$.
In the decomposition above, $\OptCenterMat$ has columns $\optcenter_{\trueclus{1}}, \dots, \optcenter_{\trueclus{\numdata}}$,
whereas $\NoiseMat$ has columns $\noisevar_1, \dots, \noisevar_{\numdata} \in \R^\dimension$, where
\[
\noisevar_i \simind \mathsf{N}(0, \Sigma_i), \quad
i = 1, \dots, \numdata.
\]

\ble
\label{lem:conc}
Let $Z \in \R^{\dimension \times \numdata}$ be a
random matrix with columns $z_i \simind \mathsf{N}(0, \Sigma_i)$,
for $i = 1, \dots, \numdata$.
Let $\lambda_{\rm max} \defn \max_{i=1, \dots, n} \opnorm{\Sigma_i^{1/2}}$.
Then
\[
\P\Big\{\opnorm{Z} \geq \lambda_{\rm max} \big(\sqrt{n} + \sqrt{d} + t \big)\Big\}
\leq \exp\Big(-\frac{t^2}{2}\Big).
\]
\ele

\begin{proof}
  We note that $\opnorm{Z} \leq \lambda_{\rm max} \opnorm{W}$ almost surely
  since we can write the columns of $Z$ as $z_i = \Sigma_i^{1/2} w_i$ in distribution,
  where $w_i \simind \mathsf{N}(0, I_d)$.
  Consequently,
  \[
  \P\Big\{\opnorm{Z} \geq \lambda_{\rm max} \big(\sqrt{n} + \sqrt{d} + t \big)\Big\}
  \leq
  \P\Big\{\opnorm{W} \leq \sqrt{n} + \sqrt{d} + t\Big\}
  \leq e^{-t^2/2},
  \]
  by standard concentration bounds for Gaussian random matrices.
\end{proof}

We record the following corollary. To do so, we first consider the following `good' event,
when the operator norm of the noise matrix $\NoiseMat$ is small enough.
\begin{equation}
  \label{concentration-event}
\concevent \defn \Big\{\opnorm{G} \leq \sqrt{2\eigmax}(\sqrt{\numdata} + \sqrt{\dimension})\Big\}
\end{equation}

\bcor
We have $\P(\concevent) \geq 1-\e^{-0.08 \numdata}$.
\ecor
\begin{proof}
  In Lemma~\ref{lem:conc}, we may set
  $Z = G$, and take $\lambda_{\rm max} \defn \sqrt{\eigmax}$.
  The result follows by taking $t = (\sqrt{2} - 1)(\sqrt{\dimension} + \sqrt{n}) \geq \sqrt{0.4 n}$.
\end{proof}

To finish the proof, we first introduce an alternative algorithm which is simpler to analyze,
which we refer to in the sequel as Algorithm~\eqref{alg:low-rank}.
\begin{algdesc}[Clustering via low-rank approximation]
\label{alg:low-rank}
\emph{Algorithm for estimating clusters from samples.}
\begin{tabbing}
  {\bf input:}\quad sample matrix $X \in \R^{\dimension \times \numdata}$, number of clusters, $\numclus$ \\[1.4ex]
  \qquad \=\ 1. \emph{Compute truncated singular value decomposition of sample matrix.}\\[1.3ex]
  \qquad \=\ \qquad \qquad
  $X_\numclus \defn \sum_{i=1}^\numclus \hat \sigma_i \hat u_i \hat v_i^\T \eqcolon \hat U \hat \Sigma \hat V^\T \in \R^{\dimension \times \numdata}$\\[1.3ex]
  \qquad \=\ 2. \emph{Solve $\numclus$-means on the columns of $Y$.} Compute a solution to
  the optimization problem\\[1.3ex]
  \qquad \=\ \qquad \qquad $\mbox{minimize} \quad \frac{1}{2n} \sum_{i=1}^\numdata \|X_i - \theta_{\kappa_i}\|_2^2$, \\[1.3ex]
  \qquad \=\ \quad with variables $\theta_1, \dots, \theta_\numclus \in \R^\dimension$ and
  $\kappa_1, \dots, \kappa_n \in [k]$.\\[1.4ex]
  {\bf output:}\quad
  estimates for labelling $\anclus{\cdot} \colon [\numdata] \to [\numclus]$,
  where $\anclus{i} \defn \kappa_i^\star$, for all $i$.
\end{tabbing}
\end{algdesc}

It turns out that there is a bijection $\phi \colon [k] \to [k]$ such that
$\anclus{i} = \phi(\predclus{i})$ for each $i \in [n]$ (for additional details,
see Lemma 4.1 of \cite{loffler2019optimality}).
Therefore, by definition of
$\ell(\cdot, \cdot)$, we have that
\[
\ell(\hat{\mathsf{c}}_{\rm spec}, \mathsf{c}^\star)
= \ell(\hat{\mathsf{c}}_{\rm appx}, \mathsf{c}^\star),
\]
and so it suffices to analyze the output of the algorithm
that simply uses the rank-$k$ approximant to $X$, algorithm~\ref{alg:low-rank}.

To do so, let us first introduce the following two sets of indices:
\[
\badindices \defn \Big\{ i \in [\numdata]
: \|\optcenter_{\trueclus{i}} - \hat \theta_{i}\|_2 \geq \Delta/2
\Big\} \qquad \mbox{and} \qquad
\goodindices \defn [n] \setminus \badindices.
\]
Above, $\hat \theta_i$ is the optimal solution $\theta_i^\star$ in step 2 of
Algorithm~\ref{alg:low-rank}. We will also use the notation $\EstCenterMat \in \R^{\dimension \times n}$
to denote the matrix which has columns $\hat \theta_i$.
Note that $\EstCenterMat$ by construction is the projection of the rank-$k$
approximant $X_k$ onto the set of matrices with at most $k$ distinct columns,
in the Frobenius norm.

We then have the following proposition.
\bpr
\label{prop:bad-ind-card}
The cardinality of the set of bad indices enjoys the following
upper bound:
\[
|\badindices| \leq \frac{128 k \opnorm{G}^2 }{\Delta^2},
\]
almost surely.
\epr
\begin{proof}
Note that
\[
\fronorm{\EstCenterMat - \TrueCenterMat}^2
=
\sum_{i=1}^n \|\optcenter_{\trueclus{i}} - \hat \theta_{i}\|_2^2
\geq |\badindices| \frac{\Delta^2}{4}.
\]
Therefore, we see that
\begin{equation}
  \label{eqn:card-bound}
  |\badindices| \leq \frac{4}{\Delta^2} \fronorm{\EstCenterMat - \TrueCenterMat}^2
\end{equation}
Note that since $\EstCenterMat, \TrueCenterMat$ are both (at most) rank-$\numclus$
matrices we evidently have
\[
\fronorm{\EstCenterMat  - \TrueCenterMat}
\leq \sqrt{2k}
\opnorm{\EstCenterMat - \TrueCenterMat}
\leq \sqrt{8k}
\opnorm{\TrueCenterMat - X_k},
\]
where in the last inequality we use the optimality
of $\EstCenterMat$ along with the triangle inequality.
Adding and subtracting $X$, and then relying on Eckart-Young, the display above implies
\[
\fronorm{\EstCenterMat  - \TrueCenterMat} \leq \sqrt{32k}
\opnorm{X - \TrueCenterMat} \leq \opnorm{G} \sqrt{32 k}.
\]
Applying the display above in the cardinality bound~\eqref{eqn:card-bound}
gives the result.
\end{proof}

Additionally, we define the following sets
\[
I_j \defn \Big\{ i \in [\numdata] :
\trueclus{i} = j,~ \|\hat \theta_i - \optcenter_{\trueclus{i}} \|_2 < \Delta/2\Big\},
\qquad
j =1, \ldots, \numclus.
\]
Clearly, we have that $\{I_j\}$ form a disjoint partition of $\goodindices$.
However, as the following result shows, these sets are also pairwise disjoint and nonempty
under the estimated clustering, $\anclus{\cdot}$.
\ble
\label{lem:disjointness}
The sets $\anclus{I_j}$ are disjoint for each $j \in [\numclus]$.
\ele
\begin{proof}
  By contradiction, suppose that $\anclus{I_{j_1}} \cap \anclus{I_{j_2}} \neq \emptyset$,
  for some $j_1 \neq j_2 \in [\numclus]$.
  This happens if and only if there exists $i_1, i_2 \in \goodindices$ and $j_0 \in [\numclus]$
  for which:
  \begin{equation}
    \label{eqn:contrad-conditions}
  \trueclus{i_1} = j_1, \quad \trueclus{i_2} = j_2, \quad \mbox{and} \quad
  j_0 = \anclus{i_1} = \anclus{i_2}.
  \end{equation}
  However, then we note that
  \begin{align*}
    \|\optcenter_{j_1} - \optcenter_{j_2}\|_2
    &\leq \|\optcenter_{\trueclus{i_1}} - \estcenter_{\anclus{i_1}}\|_2
    + \| \estcenter_{\anclus{i_1}} - \estcenter_{\anclus{i_2}}\|_2
    + \|\estcenter_{\anclus{i_2}} - \optcenter_{\trueclus{i_2}}\|_2 \tag{by triangle inequality, conditions~\eqref{eqn:contrad-conditions}} \\
    &= \|\optcenter_{\trueclus{i_1}} - \estcenter_{j_0}\|_2
    + \|\estcenter_{j_0} - \optcenter_{\trueclus{i_2}}\|_2 \tag{since $j_0 = \anclus{i_1} = \anclus{i_2}$} \\
    &< \Delta, \tag{since $i_1, i_2 \in \goodindices$}
  \end{align*}
  which contradicts the definition of $\Delta$.
\end{proof}

\section{Other deferred proofs}
\label{sec:others}
\begin{proof}[Proof of Lemma~\ref{lem:ar-decomp}]
Following the normal equations for the scalar autoregression problem
~\eqref{prob:scalar-AR},
\begin{equation}\label{eqn:decomp-of-ar-estimate}
\ARest_i = \theta_i + \EmpCov_i^{-1} \bigg\{\frac{1}{T} 
\sum_{t=1}\T (\eps_i^{(t)} + \Delta_i^{(t)}) x_i^{(t)} \bigg\}.
\end{equation}
Above we defined $\Delta_i^{(t)} \defn \sum_{j: c_j = c_i} \gamma_{ij}^\T x_j^{(t)}$.
Note that 
\[
\Delta_i^{(t)} \stackrel{\rm d}{=} \mathsf{N}\Big(0, (\tau \rho^{(t)}_i)^2\Big), 
\qquad \mbox{where} 
\quad 
\rho_i^{(t)} \defn \bigg(\sum_{j : c_j = c_i} \|x_j^{(t)}\|_2^2\bigg)^{\!\tfrac{1}{2}}.
\]
Therefore in view of~\eqref{eqn:decomp-of-ar-estimate}, we 
see that $\ARest_i$ 
is indeed a sum of Gaussians with mean $\overline{\theta}_{c_i}$ and 
covariance 
\[
\Lambda_i = \nu^2 I_d + \frac{\sigma^2}{T} \EmpCov_i^{-1} + 
\frac{\sigma^2}{T} \EmpCov_i^{-2}\EmpCov_i(\rho_i), 
\]
which proves the result. \qedhere 
\end{proof}

\begin{proof}[Proof of Lemma~\ref{lem:VAR-guarantee}]
To lighten notation, we define 
\[
Z_t \defn I_m \otimes (x^{(t)})^\T, 
\quad t \in [\numsamples].
\]
Note that $Z_t \in \R^{m \times m^2 d}$. 
It follows from the optimality condition for the 
VAR problem that we have the following decomposition: 
\[
\VARest = \bigg(\frac{1}{\numsamples} 
\sum_{t = 1}^\numsamples Z_t^\T Z_t\bigg)^{-1} 
\frac{1}{\numsamples} \sum_{t=1}^\numsamples
Z_t^\T \varepsilon^{(t)}
 = \Gamma^\star + \frac{1}{\numsamples} \sum_{t=1}^\numsamples
Z_t^\T \varepsilon^{(t)},
\]
where above we use the isotropy assumption. The decomposition above
then immediately implies that
\[
\VARest \sim \mathsf{N}(\Gamma^\star, \frac{\sigma^2}{T} I_{m^2 d}).
\]
Note that $\tfrac{T}{\sigma^2} \|(\VARest - \Gamma^\star)\|_2^2$ is 
equal in distribution to a $\chi^2$ random variable with $m^2 d$-degrees
of freedom. The result now follows by using 
standard concentration bounds for $\chi^2$-random variables.
\end{proof}

\section{Implementation Details}
\label{sec:implement}

{\bf Simulation Study.} In the simulation experiment in Figure~\ref{fig:simexp}, we generate a synthetic dataset, where there are $n$ time series grouped into $k$ clusters of equal size. In cluster $i$, the local AR parameters for every time series is centered at $\theta_{c(i)} = z_i/C_i$ where $z_i$ is generated from $\mathsf{N}(0, I_d)$ distribution and $C_i$ is a constant. The noise around $\theta_{c(i)}$ is set to have a standard deviation of 1e-2. The VAR parameters for time series in a cluster is also generated from the standard normal distribution. In other words $\gamma_{i,j} = z_{ij}/C_{ij}$ where $z_{ij} \sim \mathsf{N}(0, I_d)$ if $c(i) = c(j)$. The constants $C_i$'s and $C_{ij}$'s are chosen such that the $\ell_p$ norm of all the lag coefficients of any particular time series is one for $p =2.5$. The choice of the normalization is made to promote stationarity. We also restrict the values of the time series between $[-1, 1]$. Thus the time-series pregression is governed by the equation,

\begin{align}
	y_i^{(t)} = \min \left( 1, \max \left( \theta_{i}^\T y^{(t-d:t-1)}_{i} + \sum_{j: c(i) = c(j)} \gamma_{ij}^\T y_{j}^{(t-d:t-1)} + \epsilon_i^{(t)}, -1\right)\right).
\end{align}

{\bf Learning Algorithm.} The scalar AR algorithm is implemented using one LinearSVR~\cite{fan2008liblinear} instance per time series. Similarly the VAR models are also implemented using one LinearSVR instance per cluster. We tune the parameters of the AR model on a randomly chosen time series from the whole datasets and then held fixed for all other time series. The hyper-parameters are tuned via grid-search among the grid \texttt{\{"max\textunderscore iter": [1000], "C": np.logspace(-4, 0, 10), "epsilon": [0, 1e-1, 1e-2, 1e-3]\}}. Similarly the VAR hyper-parameters are tuned on a randomly chosen cluster among the same grid parameters as above, and then held fixed. 

\textbf{Lag Indices.} These are very important parameters for both the AR and VAR model. For hourly datasets like Electricity and Traffic, we use the lag indices \texttt{np.concatenate([np.arange(1, 25), np.arange(7*24, 8*24), np.arange(14*24, 15*24)])} i.e the last 24 hours, 24 hours on the same day last week and 24 hrs on the same day two weeks back. For the datasets in 5min intervals, like  ECG and PEMS07 we set the lag indices as \texttt{np.concatenate([np.arange(1, 12), np.arange(12, 15), np.arange(24, 27)])} i.e the last hour, the corresponding 15 min in the last hour and the corresponding 15 minutes two hours back. 

{\bf Clustering.} We cluster on the learnt AR parameters normalized to $\ell_2$ norm one. We fit a KD-Tree~\cite{bhatia2010survey} in the AR paremeter set and form a $K$-nn graph using 11 neighbors for each point. Then we use METIS graph clustering on this graph. In general we set the number of clusters to be $k= \floor{n/10}$ for a dataset with $n$ time series. We tuned the $k$ among $\{ \floor{n/5},\floor{n/10}, \floor{n/20} \}$, and $k= \floor{n/10}$ was close to optimal among the three on all 4 datasets.

{\bf Baselines.} Now we can present details about the baselines. For the first two experiments i.e the ones on ECG and PEMS07 all the baselines are in a setting identical to the ones used in~\cite{cao2020spectral}. We merely recreate the values from the table in~\cite{cao2020spectral}. It is a fair comparison as our algorithms are also evaluated on the exact same task. Similary for the experiments onTraffic and Electricity dataset the baseline numbers are taken from Table 2 in~\cite{sen2019think}. We make these choices for the ease of reproducibility, moving forward.

\begin{table}[!h]
	\centering
	\scalebox{0.9}{
		\begin{tabular}{lrrrrr}
			\toprule
			Dataset    & Num. Series ($n$) & Num. Training Points ($t$) & Forecast Horizon ($\tau$) & Rolling Windows ($n_w$) & Granularity \\
			\midrule
			Electricity & 370               & 25,968                     & 24                        & 7                    & Hourly       \\
			Traffic     & 963               & 10,392                     & 24                        & 7                    & Hourly       \\
% 			Wiki        & 115,084           & 747                        & 14                        & 4                    & Daily        \\
% 			METR-LA     & 207               & 30,846                     & 3                         & 1142                 & 5min         \\
% 			PEMS-BAY    & 325               & 46,905                     & 3                         & 1737                 & 5min         \\
			PEMS07      & 208               & 11,232                     & 3                         & 480                  & 5min         \\
			ECG         & 140               & 4,999                      & 3                         & 167                  & 5min         \\
			\bottomrule
		\end{tabular}%
	}
	\caption{\footnotesize We present the salient features of the datasets. We use the last $n_w \times \tau$ time-points to construct the test sets. Then, we perform rolling validation with a forecast horizon of $\tau$ and number of rolling windows as $n_w$.}
	\label{tab:stats}
\end{table}

\section{Metrics}
\label{sec:metrics}

The following well-known loss metrics are used in this paper. Here, $\*Y \in \R^{n'\times t'}$ represents the actual values while  $\hat{\*Y} \in \R^{n'\times t'}$ are the corresponding predictions.

{\bf (i) WAPE: } Weighted Absolute Percent Error is defined as follows,
\begin{align}
\label{eq:wape}
\_L_w(\hat{\*Y},\*Y) = \frac{\sum_{i = 1}^{n'}\sum_{j=1}^{t'} |Y_{ij} - \hat{Y}_{ij}|}{\sum_{i = 1}^{n'}\sum_{j=1}^{t'} |Y_{ij} |}.
\end{align}

{\bf (ii) MAPE: } Mean Absolute Percent Error is defined as follows,
\begin{align}
\label{eq:mape}
\_L_m(\hat{\*Y},\*Y) = \frac{1}{Z_0}\sum_{i = 1}^{n'}\sum_{j=1}^{t'} \frac{|Y_{ij} - \hat{Y}_{ij}|}{ |Y_{ij} |} \mathbf{1}\{|Y_{ij}| > 0\},
\end{align}
where $Z_0 = \sum_{i = 1}^{n'}\sum_{j=1}^{t'} \mathbf{1}\{|Y_{ij}| > 0\}$.

{\bf (iii) SMAPE: } Symmetric Mean Absolute Percent Error is defined as follows,
\begin{align}
\label{eq:smape}
\_L_s(\hat{\*Y},\*Y) = \frac{1}{Z_0}\sum_{i = 1}^{n'}\sum_{j=1}^{t'} \frac{2|Y_{ij} - \hat{Y}_{ij}|}{ |Y_{ij}| + |\hat{Y}_{ij}|} \mathbf{1}\{|Y_{ij}| + |\hat{Y}_{ij}| > 0\},
\end{align}
where $Z_0 = \sum_{i = 1}^{n'}\sum_{j=1}^{t'} \mathbf{1}\{Y_{ij}| + |\hat{Y}_{ij}| > 0\}$.

{\bf (iv) MAE: } Mean Absolute Error is defined as follows,
\begin{align}
\label{eq:wape}
\_L_{mae}(\hat{\*Y},\*Y) = \frac{1}{n't'}\sum_{i = 1}^{n'}\sum_{j=1}^{t'} |Y_{ij} - \hat{Y}_{ij}|
\end{align}

{\bf (iv) RMSE: } Root Mean Squared Error is defined as follows,
\begin{align}
\label{eq:wape}
\_L_{rmse}(\hat{\*Y},\*Y) = \sqrt{\frac{1}{n't'}\sum_{i = 1}^{n'}\sum_{j=1}^{t'} (Y_{ij} - \hat{Y}_{ij})^2}.
\end{align}

\subsection{Linear Generative Model and Analogy to Mixed Linear Regression} 
\label{sec:lgn}

For the sake of completeness, we define a linear generative model for our time-series setting under which we expect our algorithm to recover the true parameters. In the second half of this section, we will relate this model to the simplified mixed linear regression problem, under which we formally prove our guarantees.

We will consider an auto-regressive lag of $d$.  That is, any data point at a given time can depend only on quantities no more than $d$ time steps in the past. We consider a dataset with $n$ time series and $T$ time points. The following process selects the local autoregressive parameters for each time series $i \in [n]$:
\begin{subequations}
	\begin{align}
		\trueclus{i} &\sim \mathsf{Categorical}(\pi_1, \dots, \pi_k), \quad \mbox{and} \quad \\ 
		\theta_i \mid \trueclus{i} &\sim \mathsf{N}_{\dimension}(\overline{\theta}_{\trueclus{i}}, \nu^2 I_d),
	\end{align}
\end{subequations}
where $\mathsf{Categorical}$ and $\mathsf{N}_{\dimension}$ denote the Categorical and $\dimension$-dimensional Gaussian distributions respectively. 

This means that the time series $i$ is first randomly assigned into one of $k$ clusters $\trueclus{i}$. Based on the cluster, the local autoregressive parameters of the time series are drawn from the $\trueclus{i}$-th component of a Gaussian Mixture model in $\R^{d}$. The evolution of the time series is then given as follows:
\begin{equation}
	\label{eq:spec}
	x_i^{(s)} = \theta_i^\T x_i^{(s-d:s-1)} + \sum_{\substack{j \neq i\\ \trueclus{i} = \trueclus{j}}} \gamma_{ij}^\T x_j^{(s-d:s-1)} + \epsilon_i^{(s)},
\end{equation}
for $s \in \{d+1, \cdots, T\}$. Here, $\gamma_{ij} \in \R^{d}$ are parameters that govern how time series $i$ depends on the past of time series $j$, only if both of them belong to the same cluster. This essentially implies that apart from the dependence on its own past, a time series can depend on the past of all other time series in the same cluster. Moreover the clustering is defined by closeness with respect to the local autoregressive parameters. From Equation~\eqref{eq:spec}, we see that the values of a time series do not depend on those from clusters that the corresponding time series does not belong to. This is a reasonable assumption in real-world multivariate time-series settings, where each time series would only depend on a few related time series.

{\bf Relation to MLR:} The mixed linear regression problem is an i.i.d analogue of the linear generative model presented above. Each time-series can be thought of as a scalar regression problem, where each data-point in the MLR problem corresponds to the target being the future time-point and the covariates being the past $d$ time-points, i.e
\begin{align}
(x_i^{(j)}, y_i^{(j)}) \sim (x^{(s_j-d:s_j-1)}_{i}, x_{i}^{(s_j)}).
\end{align}
where $s_j$ corresponds to the blocks separated by the mixing time, defined in Section~\ref{sec:framework}. Time-series that come from the same cluster would have $\theta_i$'s from the same component of the mixture in MLR model defined in Section~\ref{sec:theory}. The time-series in the same cluster will have further VAR dependencies on each other as captured by $\gamma$ coefficients in the MLR model.

\clearpage

\end{document}